\documentclass[sigconf]{acmart}

\AtBeginDocument{%
  }

\setcopyright{none}
\renewcommand\footnotetextcopyrightpermission[1]{} 
\makeatletter
\def\@copyrightspace{\relax}
\makeatother

\usepackage{multirow}
\usepackage{subcaption,booktabs}

\begin{document}

\title{ViTGuard: Attention-aware Detection against Adversarial Examples for Vision Transformer}

\author{Shihua Sun}
\affiliation{%
  \institution{Virginia Tech}
  \city{Arlington}
  \state{VA}
  \country{USA}
}
\email{shihuas@vt.edu}

\author{Kenechukwu Nwodo}
\affiliation{%
  \institution{Virginia Tech}
  \city{Arlington}
  \state{VA}
  \country{USA}
}
\email{nwodok@vt.edu}

\author{Shridatt Sugrim}
\affiliation{%
  \institution{Kryptowire Labs}
  \city{McLean}
  \state{VA}
  \country{USA}
}
\email{ssugrim@kryptowire.com}

\author{Angelos Stavrou}
\affiliation{%
  \institution{Virginia Tech}
  \city{Arlington}
  \state{VA}
  \country{USA}
}
\email{angelos@vt.edu}

\author{Haining Wang}
\affiliation{%
  \institution{Virginia Tech}
  \city{Arlington}
  \state{VA}
  \country{USA}
}
\email{hnw@vt.edu}


\renewcommand{\shortauthors}{S. Sun et al.}

\begin{abstract}
The use of transformers for vision tasks has 
challenged the traditional dominant role of convolutional neural networks (CNN) in computer vision (CV).
For image classification tasks, Vision Transformer (ViT) effectively establishes spatial relationships between patches within images, directing attention to important areas for accurate predictions.
However, similar to CNNs, ViTs are vulnerable to adversarial attacks, which mislead the image classifier into making incorrect decisions on images with carefully designed perturbations. Moreover, adversarial patch attacks, which introduce arbitrary perturbations within a small area (usually less than 3\% of pixels), pose a more serious threat to ViTs. 
Even worse, traditional detection methods, originally designed for CNN models, are impractical or suffer significant performance degradation when applied to ViTs, and they generally overlook patch attacks.

In this paper, we propose ViTGuard as a general detection method for defending ViT models against adversarial attacks, including typical attacks where perturbations spread over the entire input ($L_p$ norm attacks\footnote{In $L_p$ norm attacks, the perturbation is bounded by $L_p$ norm distance metrics, namely $L_1$, $L_2$, and $L_\infty$.}) and patch attacks.
ViTGuard uses a Masked Autoencoder (MAE) model to recover randomly masked patches from the unmasked regions, providing a flexible image reconstruction strategy.
Then, threshold-based detectors leverage distinctive ViT features, including attention maps and classification (CLS) token representations, to distinguish between normal and adversarial samples. The MAE model does not involve any adversarial samples during training, ensuring the effectiveness of our detectors against unseen attacks. 
ViTGuard is compared with seven existing detection methods under nine attacks across three datasets with different sizes. The evaluation results show the superiority of ViTGuard over existing detectors. Finally, considering the potential detection evasion, we further
demonstrate ViTGuard's robustness against adaptive attacks for evasion.
\end{abstract}

\maketitle

\section{Introduction}
The attention-based transformer models~\cite{attention_all} have emerged as the foundational cornerstone across a diverse spectrum of applications in the field of natural language processing (NLP), such as Pathways Language Model (PaLM)~\cite{palm} 
developed by Google and Generative Pre-trained Transformer (GPT)~\cite{gpt} developed by OpenAI. Motivated by the applications of transformers in the realm of NLP, researchers have extended their utilization to the computer vision (CV) domain, which was traditionally dominated by Convolutional Neural Networks (CNN). Transformers have demonstrated notable success in diverse CV applications, such as image classification~\cite{vit, deit}, segmentation~\cite{segmentation}, image generation~\cite{igpt}, and so on. For instance, for image classification tasks, vision transformers (ViT) have proven their ability to achieve state-of-the-art or superior accuracy on large datasets 
compared to CNNs, while requiring reduced computational resources~\cite{vit, r_learner}. 
However, despite their empirically demonstrated robustness compared to CNNs~\cite{reveal_robust, Intriguing_properties, robustart_2022}, ViTs are still prone to adversarial (evasion) attacks~\cite{vit_cnn_fair, understanding_robustness, robustness_vit, robustness_vit2}, resulting in the misclassification of inputs with visually imperceptible perturbations. Moreover, it has been proven that ViTs are more vulnerable to patch attacks than CNNs by misdirecting attention to adversarial patches. 
While a majority of prior work focused on $L_p$ norm attacks, recent research has identified patch attacks as a more practical problem in real-world scenarios~\cite{adv_patch, PatchCensor, Perceptual-Sensitive, bias-based}, such as attaching an adversarial patch to a stop sign.
This vulnerability limits the adaptation of transformers in security-sensitive applications, such as face recognition and self-driving cars. 


Countermeasures against adversarial examples in the context of CNN models have been extensively studied. These defenses can be broadly divided into two categories: (1) detecting adversarial examples and removing them during testing~\cite{KDBU, HGD, nic, Lid, magnet, DNR, AmI_interpretability, Feature_squeezing}, and (2) enhancing the model's robustness to thwart adversarial attempts~\cite{FGSM,  random_2, defensive_distillation,  blackbox_1, defense-gan, random_1}. In terms of ViT models, current efforts mainly focus on enhancing ViTs' robustness via data augmentation~\cite{defense1_certifiable}, adversarial training~\cite{vit_adv_training}, attention mechanism modification~\cite{smooth_attn_2022, RVT2022, FAN2022}, or ensemble strategies~\cite{robustness_vit}, these methods typically require architecture modifications or retraining. Moreover, applying existing detection methods designed for CNNs to ViTs is either infeasible or results in severely degraded detection performance given the distinct architectures. Additionally, CNN detection methods typically ignore patch attacks that are more prevalent against ViTs. Therefore, a general mechanism for detecting both $L_p$ norm and patch attacks, tailored for the ViT architecture and capable of being applied in tandem with the ViT model without the need to modify or retrain it, is in great demand but still missing.

\begin{figure}[t]
\centerline{\includegraphics[width=\linewidth, trim =0.7cm 0.4cm 15cm 7.5cm, clip]{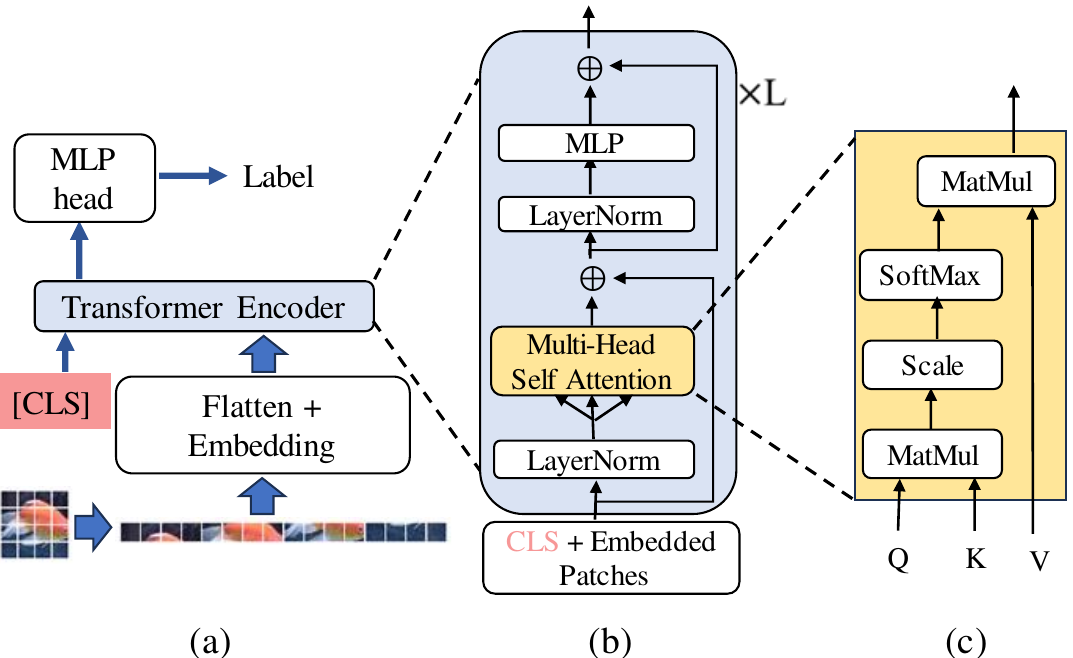}}
\caption{Vision Transformer architecture: (a) the main structure of ViT, (b) the transformer block, and (c) the self-attention mechanism. }
\label{fig:vit_arch}
\vspace{-0.1in}
\end{figure}

The ViT model~\cite{vit} has a significantly different architecture from CNNs. It begins by dividing input images into patches, subsequently sending patch embeddings to the transformer encoder, which serves as the backbone of ViT. As depicted in Figure~\ref{fig:vit_arch}, instead of being built upon convolutional layers, the transformer encoder relies on multi-head self-attention to establish a global correlation (attention map) between patches. This design allows ViT to focus on the most important areas with high attention scores for accurate classification. 
Additionally, a distinctive feature of ViT is the appending of a classification (CLS) token to input patches (see Figure~\ref{fig:vit_arch}(a)). 
As the CLS token and embedded patches pass through the transformer layers shown in Figure~\ref{fig:vit_arch}(b), the learnable CLS token is updated with information from other patches. A final decision can be made using only the CLS token. 
Therefore, given this revolutionary attention-based transformer architecture, we aim to design a detection mechanism that leverages the unique intrinsic features within the ViT model
to identify adversarial samples. 

In this paper, we propose a novel detection method, named ViTGuard, for ViT models to counter adversarial examples by utilizing the features of \textit{attention map} and \textit{CLS representation} (i.e., the latent representation of the CLS token). The attention map captures the global correlation among different patches, while the CLS representation encapsulates crucial information learned from image patches for classification. Our key idea is to compare attention maps and CLS representations 
of the reconstructed image by Masked Autoencoder (MAE)~\cite{masked_autoencoder} with those of the original image (i.e., the test example before reconstruction). 
MAE is a self-supervised learning approach that randomly masks out a portion of input images and then predicts the masked information from the unmasked regions. It employs an encoder-decoder design similar to traditional autoencoders~\cite{autoencoder_1}, but it is constructed using self-attention-based transformer blocks. Compared to other adversarial image recovery methods, such as denoising autoencoders (DAE) and generative adversarial networks (GAN), MAE has three main advantages: (1) it randomly reconstructs a part of the input image, resulting in lower reconstruction loss for normal images,  (2) its attention-based architecture accurately recovers the relationships between image patches, and (3) it provides a more flexible approach by employing various image recovery strategies and masking ratios to balance detection effectiveness and efficiency.

\begin{figure}[t]
\centerline{\includegraphics[width=\linewidth, trim =2.5cm 7.7cm 10cm 6.7cm, clip]{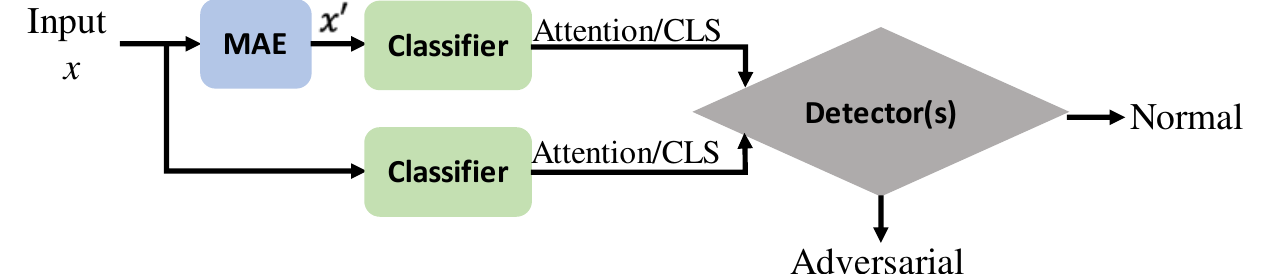}}
\caption{The workflow of ViTGuard. $x$ and $x^\prime$ represent the original and reconstructed inputs, respectively. ViTGuard is only applied in the test phase.}
\label{fig:workflow}
\vspace{-0.1in}
\end{figure}

A larger discrepancy in attention maps and CLS representations implies a greater likelihood that the input is adversarial. 
As shown in Figure~\ref{fig:workflow}, the workflow of our detection is: (1) using the MAE model to reconstruct inputs that are partially masked; (2) forwarding both the original and reconstructed images to the target ViT model and extracting their corresponding attention maps and CLS representations from a specified transformer layer;
(3) employing two detectors to compute the $L_2$ distance for both extracted features between the original and reconstructed images. Eventually, for each detector, if the $L_2$ distance is larger than a tuneable threshold, the input is identified as adversarial; otherwise, it is identified as normal. 
Thresholds are set based on acceptable false positive rates (FPR) of 1\% and 5\% in this study.

 Our detectors do not need prior knowledge of attacks launched in the testing phase, and they can be applied in conjunction with different ViT models. Additionally, by automating the intermediate layer selection used to extract attention maps and CLS representations, our detection method is easy to use and does not need any complex parameter adjustments. Our detectors are evaluated using ViT models with patch sizes of 16$\times$16 and 32$\times$32. The evaluations are conducted on three datasets: CIFAR-10, CIFAR-100, and Tiny-ImageNet. For our detectors, the Area Under the ROC Curve (AUC) scores (ranging from 0 to 1), approach 1 or reach 1 with the superior detector. 
Overall, our major contributions are summarized as follows.

\noindent
  (1) 
We propose ViTGuard, an adversarial sample detection mechanism for ViT models, which is effective for detecting both $L_p$ norm attacks and patch attacks.
 Our method's effectiveness is validated against a broad spectrum of adversarial attacks, including four typical white-box $L_p$ norm attacks~\cite{CW, APGD, FGSM, PGD}, two ViT-specific white-box patch attacks~\cite{PatchFool, attn_fool}, a transfer-based attack~\cite{SGM} using a CNN source model, and two transfer-based attacks~\cite{transfer_se_tr} employing a ViT source model.

\noindent
 (2) In the absence of an advanced detection method capable of addressing both types of attacks for ViTs, we compare our approach with six typical detection methods designed for $L_p$ norm attacks, as well as a state-of-the-art adversarial patch detection method.
 

\noindent
(3) By studying intermediate layer choices for feature extraction, 
we reveal that the deep layer representations commonly utilized in detection for CNN are not necessarily effective for ViT models.

\noindent
(4) We demonstrate ViTs' robustness against possible evasion by
developing an adaptive attack, assuming that 
attackers are aware of the detection and attempt to evade it.

\section{Background \& Related Works}
In this section, we present the background on the ViT architecture and adversarial attacks, and we also survey existing defenses.

\subsection{Vision Transformer}
Similar to typical machine learning (ML) models for image classification, ViT~\cite{vit} provides a function $f: \boldsymbol{x} \rightarrow y $ to map from the input to its label. The architecture of ViT~\cite{vit, attention_all} is illustrated in Figure~\ref{fig:vit_arch}. The workflow of ViT involves: (1) splitting the input image into patches, flattening the patches, and applying linear embeddings to them; (2) appending a CLS token at the beginning of the patch embeddings; (3) adding position embeddings; (4) feeding sequences into the transformer encoder; (5) forwarding the representation of the CLS token to a multilayer perceptron (MLP) head for final classification. In addition, variants of ViT have emerged with the development of transformers in the CV domain, such as Data-efficient image Transformer (DeiT)~\cite{deit}, Class-Attention in Image Transformers (CaiT)~\cite{cait}, Shifted Windows (Swin) Transformer~\cite{swin}, Tokens-to-Token ViT (T2T-ViT)~\cite{T2T-ViT}, and Pyramid Vision Transformer (PVT)~\cite{PVT}.

There are several important components and mechanisms within the ViT model that distinguish it from conventional Deep Neural Network (DNN) models. The Self-Attention Mechanism captures the relationship between tokens by weighting the importance of each token relative to others. The attention mechanism is shown in Figure~\ref{fig:vit_arch} (c), where Q, K, and V represent the Query, Key, and Value vectors, respectively, which are projected by linear transformations of the input embeddings. The Multi-Head Self Attention (MSA) Layer gets richer representations from different feature subspaces. As shown in Figure~\ref{fig:vit_arch} (b), each transformer block consists of two sub-layers that incorporate an MSA and an MLP, respectively. Additionally, each sub-layer has a residual connection around it. The CLS token, inherited from NLP transformers, is appended to the patches before being fed into the encoder and is ultimately utilized in the MLP head for the final classification. 


\subsection{Adversarial Attacks on CNNs}
Adversarial attacks fall into two categories based on the attacker's knowledge of the classifier's architecture and parameters: white-box (complete knowledge) and black-box (no knowledge). Additionally, according to the perturbed areas, the attacks are categorized into 
$L_p$ norm attacks and patch attacks.

\textit{White-Box Attacks.} The Fast Gradient Sign Method (FGSM)~\cite{FGSM}, Projected Gradient Descent (PGD) attack ~\cite{PGD} and Auto PGD (APGD) attack \cite{APGD} add perturbations into images with the objective of maximizing the loss function. The Jacobian-based Saliency Map Attack (JSMA)\cite{JSMA} selectively alters salient pixels based on an adversarial saliency map. DeepFool\cite{deepfool} iteratively adjusts samples to cross the decision boundary. The Carlini and Wagner (CW) attack~\cite{CW} formulates the adversarial attack as an optimization problem, with an objective function comprising two terms: the perturbation magnitude and a criterion indicating misclassification of the adversarial sample. Furthermore, Liu et al.\cite{bias-based} and Karmon et al.\cite{lavan} investigated patch attacks on CNN models, creating transferable adversarial patches that occupy just 2\% of the image area.

\textit{Black-Box Attacks.}
 Black-box adversarial attacks can be categorized into transfer-based~\cite{blackbox_3_drmi, blackbox_2, blackbox_1} and query-based~\cite{query_2_square, query_3, boundary_attack, query_1_rays, zoo_blackbox, query_4_qeba}.
Transfer-based attacks leverage adversarial transferability to generate adversarial examples by using a substitute target model. \cite{trans_momentum, SGM, TIM, SIM, DIM} further improved the transferability of adversarial examples by modifying gradient calculations or increasing the diversity of input patterns.
Query-based attacks, on the other hand, rely on the target model's outputs. By iteratively sending queries and analyzing the feedback from the model, the query-based attacker can generate adversarial examples based on the estimated model information, such as gradients. 

\subsection{Adversarial Attack on ViTs}
ViTs remain vulnerable to the aforementioned adversarial attacks originally designed for CNNs~\cite{robustness_vit, robustness_vit2}, which we will not detail again here. Instead, we describe attacks that have gained significant attention within the domain of transformer models.

\textit{White-Box Attacks.}
Fu et al.\cite{PatchFool} and Lovisotto et al.\cite{attn_fool} proposed the patch attacks Patch-Fool and Attention-Fool, respectively. Both perturb a small number of patches using attention-aware losses, with Patch-Fool utilizing post-softmax attention scores and Attention-Fool employing pre-softmax attention scores.

\textit{Black-Box Attacks.}
To improve adversarial transferability, Naseer et al.~\cite{transfer_se_tr} introduced two methods, Self-Ensemble (SE) and Token Refinement (TR), for the unique architecture of ViTs. SE is designed to deceive a model ensemble, where each transformer block is followed by the final MLP head. TR mitigates feature misalignment by inserting intermediate layers into the model ensemble.
Wei et al.~\cite{pna_patchout_2022} proposed ignoring the gradients of attention and enhancing input diversity by randomly sampling perturbed patches to improve adversarial transferability. Zhang et al.\cite{transfer_TGR2023} proposed the Token Gradient Regularization (TGR) method, which selects tokens with extreme gradients and eliminates these gradients to reduce gradient variance.
Regarding query-based attacks, Shi et al.~\cite{patch_removal} observed the varying noise sensitivities among patches and developed a technique called Patch-wise Adversarial Removal (PAR), which selectively reduces noise across these patches.

\subsection{Existing Detection and Defense Methods for CNNs} 
We categorize current detection methods into supervised and unsupervised detection based on whether adversarial samples are involved in the training dataset for detectors. In this paper, our goal is to develop an unsupervised approach. It is important to note that detection methods for CNNs are primarily focused on $L_p$ norm attacks and do not address patch attacks.

\textit{Supervised detection.} 
Feinman et al.\cite{KDBU} and Ma et al.\cite{Lid} proposed different methods, such as Kernel Density (KD) estimates, Bayesian Uncertainty (BU) estimates, and Local Intrinsic Dimensionality (LID), to characterize the properties of adversarial and normal samples. These methods have shown limited effectiveness against advanced adversarial attacks~\cite{CW}.
Yang et al.~\cite{ml-loo} introduced a method called ML-LOO, which uses variations in feature contributions to detect adversarial examples. However, scaling ML-LOO to datasets with a large feature space is impractical because images with each feature masked are treated as separate inputs for detection. Given that pre-trained ViT models typically resize inputs to $224\times224\times3$ or $384\times384\times3$, ML-LOO is not suitable for protecting ViTs.
Moreover, Kherchouche et al.~\cite{MSCN} used the mean subtracted contrast normalized (MSCN) coefficients to extract feature statistics directly from input samples. 
Overall, the drawback of these methods is their reliance on involving adversarial samples during the detectors' training process.

\textit{Unsupervised detection.} 
Xu et al.~\cite{Feature_squeezing} proposed to use \textbf{F}eature \textbf{S}queezing (FS)~\cite{Feature_squeezing} techniques to compress the unnecessarily large input feature spaces. 
However, in practice, we observe that the feature squeezers require complex parameter adjustments for different datasets.
Meng et al.\cite{magnet} and Liao et al.\cite{HGD} utilized CNN-based denoising autoencoders to map adversarial examples onto the normal manifold and used reconstruction errors and deep layer outputs to detect adversarial samples. 
Instead of relying solely on the output of a single layer, \textit{DNR} (Deep Neural Rejection)~\cite{DNR} leverages representations from three deep layers to construct Support Vector Machines (SVM) classifiers. 
Ma et al. proposed NIC~\cite{nic} to leverage two channels: the activation value distribution channel and the provenance channel. The former deals with activation instability in a single layer due to input perturbations, while the latter focuses on instability in the activation output of the next layer caused by perturbations in the current layer's activation. However, ViT does not follow the layer-by-layer architecture, where each layer consists of a CNN layer followed by an activation function. Thus, the properties utilized by NIC, i.e., activation instabilities in each layer and between layers, do not exist in ViT, making this method impractical for application to ViT.

\textit{Other defenses aiming to improve robustness.}
In addition to the aforementioned detection mechanisms, various defenses have been developed to enhance the robustness of target models against adversarial attacks, ensuring correct classifications on adversarial examples. (1) For $L_p$ norm attacks, adversarial training~\cite{FGSM, adversarial_training3, adversarial_training2} involves adversarial examples with true labels into the training dataset of the target classifier. The effectiveness of this approach is significantly compromised if adversarial examples are not generated from the identical adversarial attack conducted in the testing phase. Another defense mechanism is to introduce randomness~\cite{random_1, random_2} to the target model by adding noise to the training data and intermediate layers' outputs. Gradient masking~\cite{blackbox_1} mitigates attacks by concealing useful gradients. Defensive distillation~\cite{defensive_distillation} trains a smoother model with soft labels. Defense-GAN~\cite{defense-gan} and DiffPure~\cite{diffpure} utilize GAN and diffusion models, respectively, to map inputs onto the manifold of clean data, effectively removing adversarial perturbations.  
However, diffusion models introduce a substantial computational load, as they require hundreds of iterations to reconstruct an image~\cite{diffpure}. Consequently, defenses based on diffusion models are impractical and have been excluded from our analysis.
 (2) For patch attacks, the Clipped BagNet (CBN)~\cite{bagnet} method and the Derandomized Smoothing (DS)~\cite{derandomized_smoothing} method iteratively process segments of the input sample through the classifier and determine the final outcome based on the consensus of the classification results.
Xiang et al.~\cite{PatchFool} utilized CNN models with small receptive fields to limit the influence of adversarial patches and designed a robust masking method to eliminate adversarial features.


\subsection{Existing Detection and Defense Methods for ViTs} 

\textit{Detection for ViTs.}
Currently, there is a lack of ViT-specific detection methods designed for $L_p$ norm attacks.
For patch attacks, Liu et al.~\cite{armro} observed that adversarial patches typically activate in the middle transformer layers and proposed AbnoRmality-Masking RObust (ARMRO) for detecting suspicious patches with high attention scores in the activate stage, subsequently masking these patches from the input.  Although ARMRO was designed to identify adversarial patches rather than adversarial examples, we still use it as a baseline to demonstrate the capability of our detectors against patch attacks, due to the absence of alternative methods.

\textit{Other defenses aiming to improve robustness.} 
(1) For $L_p$ norm attacks,
Gu et al.~\cite{smooth_attn_2022} applied a softmax function with a temperature parameter to smooth attention scores in transformer blocks. 
Mo et al.~\cite{vit_adv_training} proposed two warm-up strategies: randomly dropping the gradient flow in some attention blocks and masking perturbations in certain input patches.
Mao et al.~\cite{RVT2022} created a new ViT model termed Robust Vision Transformer (RVT) by combining robust key components within ViTs.
Zhou et al.~\cite{FAN2022} introduced a channel processing module within the transformer block, enhancing ViTs' robustness by intelligently aggregating embeddings from various tokens.
(2) For patch attacks,
Chen et al.\cite{defense1_certifiable} and Salman et al.\cite{certified_2022} adapted the DS method for ViT models, determining the classification result based on the majority vote. Guo et al.~\cite{RSPC} stabilized attentions by training ViT models using images altered with adversarial patches.
Overall, these methods either involve modifications to the ViT model architecture and attention mechanisms, require model retraining, or impose substantial overhead to enhance input diversity, all of which impede their deployment in practical applications.

\textcolor{blue}{}
\section{Threat Model and ViT Robustness}
In this section, we first present the threat model and then experimentally illustrate the vulnerability of ViT models to various types of adversarial attacks.

\noindent
\textbf{Threat Model:} 
For a thorough evaluation of robustness, we consider attackers with varying levels of knowledge about target models. (1) White-box: attackers have full access to target models, including the model architecture and weights; (2) Black-box: attackers lack access to target models. While the white-box setting represents attackers with strong capabilities, the black-box assumption reflects a more realistic setting. We assume that white-box attackers' training process is not influenced by our detection scheme. We also assume that
attackers only aim to mislead target models during the testing phase but do not disturb the training process. Furthermore, in Section~\ref{sec:adaptive_attacks}, it is assumed that the attacker has complete prior knowledge of the detection mechanism and adaptively generates adversarial samples.


\noindent
\textbf{Target Classifiers:}  We use ViT-Base models as the target models, which consist of 12 transformer blocks and 12 attention heads. We utilize the pre-trained ViT models available on Hugging Face, initially trained on ImageNet-21k~\cite{imagenet}, and then fine-tune their MLP heads for downstream tasks. To fit the data into the pre-trained model, the images are resized to 224 $\times$ 224 and then divided into patches with sizes of 16 $\times$ 16 and 32 $\times$ 32. The models with patch sizes of 16 and 32 are referred to as ViT-16 and ViT-32, respectively, throughout the rest of the paper. 
During the fine-tuning process, the transformer blocks are frozen, while the MLP head is trained for 50 epochs using a learning rate of $10^{-4}$ with the Adam optimizer.

\noindent
\textbf{Adversarial Attacks:} In this paper,  we assess the performance of target models under a broad range of attacks.
For white-box attacks, FGSM~\cite{FGSM}, PGD~\cite{PGD}, and APGD~\cite{APGD} are gradient-based methods, while CW~\cite{CW} leverages an optimization-based methodology. Furthermore, Patch-Fool~\cite{PatchFool} and Attention-Fool~\cite{attn_fool} are attention-aware patch attacks explicitly crafted for ViT models. For black-box attacks, we employ three transfer-based attacks: SGM~\cite{SGM}, SE~\cite{transfer_se_tr}, and TR~\cite{transfer_se_tr}. SGM generates adversarial examples by applying PGD to a ResNet surrogate model, while SE and TR use a ViT model, DeiT-Tiny-Patch16-224~\cite{deit-t}, as the source model. The attack parameters are detailed in Table~\ref{tab:attack_param} within Appendix~\ref{appendix_attack}. 
Specifically, using the $L_{\infty}$ metric, the FGSM, PGD, APGD, SGM, SE, and TR attacks apply a maximum perturbation of 0.03 for CIFAR-10 and CIFAR-100, and 0.06 for Tiny-ImageNet. The perturbation magnitudes of the CW attack, using the $L_2$ metric, are calculated as 1.37, 2.56, and 2.46 for the ViT-16 model, and 2.11, 2.28, and 2.55 for the ViT-32 model, corresponding to CIFAR-10, CIFAR-100, and Tiny-ImageNet, respectively. 
For patch attacks, namely Patch-Fool and Attention-Fool, only four patches for ViT-16 (196 patches) and one patch for ViT-32 (49 patches) are perturbed, with no constraints on the perturbations within the patches. The adversarial examples generated by these attacks are visualized in Figure~\ref{fig:vis_inputs}(a) within Appendix~\ref{appendix:vis_inputs}.

\noindent
\textbf{Robustness Analysis:} Table~\ref{tab:acc_under_attacks} lists the target classifiers' classification accuracy in both attack and non-attack scenarios with three image datasets, including CIFAR-10~\cite{cifar10}, CIFAR-100~\cite{cifar10}, and Tiny-ImageNet~\cite{tiny-imagenet}. We can see that the accuracy of ViT models is significantly affected by adversarial examples. Notably, patch attacks lead to considerable degradation in classification accuracy by perturbing only around 2\% of the patches.
Consequently, it is essential to design a general detection mechanism to safeguard ViT models from adversarial samples.

\begin{table}[]
\caption{Classification accuracy of target models under various adversarial attacks.}
\label{tab:acc_under_attacks}
\resizebox{\columnwidth}{!}{
\begin{tabular}{|c|cc|cc|cc|}
\hline
Dataset    & \multicolumn{2}{c|}{CIFAR-10}    & \multicolumn{2}{c|}{CIFAR-100}       & \multicolumn{2}{c|}{Tiny-ImageNet}   \\ \hline
Model      & \multicolumn{1}{c|}{ViT-16} & ViT-32 & \multicolumn{1}{c|}{ViT-16} & ViT-32 & \multicolumn{1}{c|}{ViT-16} & ViT-32 \\ \hline
No attack  & \multicolumn{1}{c|}{95.40\%}       &96.30\%    &\multicolumn{1}{c|}{85.87\%}     &85.44\%  &\multicolumn{1}{c|}{85.03\%}       &84.74\%        \\ \hline
FGSM       & \multicolumn{1}{c|}{46.71\%}       & 40.00\%  &\multicolumn{1}{c|}{22.29\%}  &20.50\%      & \multicolumn{1}{c|}{19.68\%}       &15.25\%        \\ \hline
PGD        & \multicolumn{1}{c|}{0.05\%}       &0.11\%   &\multicolumn{1}{c|}{0.01\%} & 0.06\%      & \multicolumn{1}{c|}{0.00\%}       &0.00\%        \\ \hline
APGD       & \multicolumn{1}{c|}{0.02\%}       &0.03\%    &\multicolumn{1}{c|}{0.00\%} &0.01\%    & \multicolumn{1}{c|}{0.00\%}       &0.00\%        \\ \hline
CW         & \multicolumn{1}{c|}{0.16\%}       &0.08\%    &\multicolumn{1}{c|}{0.00\%} &0.08\%    & \multicolumn{1}{c|}{0.04\%}       &0.28\%        \\ \hline
Patch-Fool & \multicolumn{1}{c|}{28.96\%}       &31.94\%   &\multicolumn{1}{c|}{7.14\%} &15.20\%     & \multicolumn{1}{c|}{15.09\%}       &20.95\%        \\ \hline
Attention-Fool & \multicolumn{1}{c|}{25.52\%}       &30.14\%   &\multicolumn{1}{c|}{5.68\%} &13.64\%     & \multicolumn{1}{c|}{13.28\%}       &18.84\%        \\ \hline
SGM & \multicolumn{1}{c|}{34.10\%}       &43.50\%   &\multicolumn{1}{c|}{38.30\%} & 46.30\%     & \multicolumn{1}{c|}{34.23\%}       & 37.69\%        \\ \hline
SE & \multicolumn{1}{c|}{63.07\%}       &87.99\%   &\multicolumn{1}{c|}{59.71\%} & 75.25\%     & \multicolumn{1}{c|}{51.59\%}       & 71.42\%        \\ \hline
TR & \multicolumn{1}{c|}{42.46\%}       &82.55\%   &\multicolumn{1}{c|}{35.39\%} & 62.15\%     & \multicolumn{1}{c|}{22.74\%}       & 53.11\%        \\ \hline
\end{tabular} }
\end{table}

\section{Framework} 

The workflow of our detection method is illustrated in Figure~\ref{fig:workflow}. In the testing phase, given an input image, the detection process of our
detectors is as follows:
(1) image reconstruction (Section~\ref{sec:intro_autoencoder}), (2) extraction of attention maps or CLS representations for the original and reconstructed images (Sections~\ref{detector-1} and~\ref{detector-2}), and (3) calculation of the distance between their attention maps or CLS representations (Sections~\ref{detector-1} and~\ref{detector-2}). If the distance exceeds a predefined threshold, the input is identified as adversarial and excluded from the testing phase. In Section~\ref{sec:strategy_joint_detectors}, we combine two individual detectors to enhance detection performance.






\subsection{Input Reconstruction} \label{sec:intro_autoencoder}

\begin{figure}[t]
\centerline{\includegraphics[width=\linewidth, trim =0.5cm 6.25cm 2.9cm 6.7cm, clip]{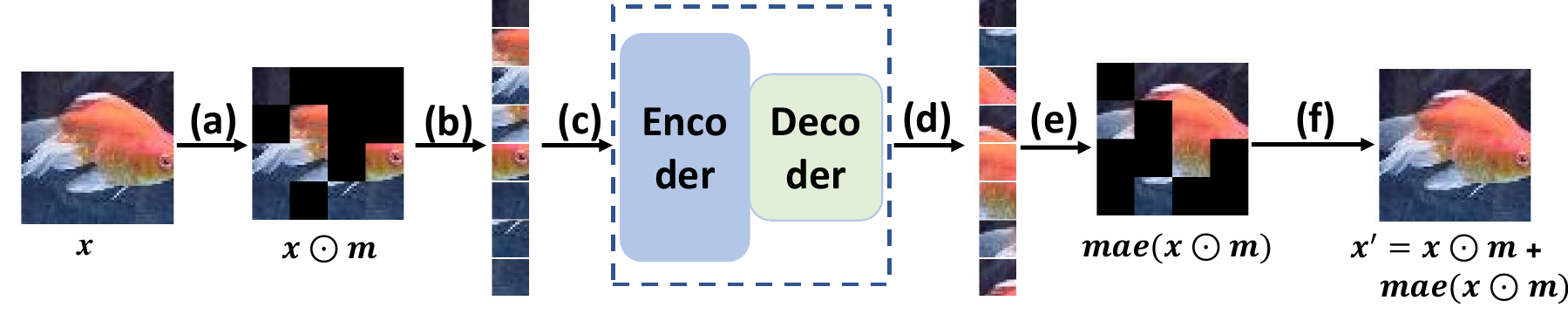}}
\caption{The architecture and workflow of MAE. Note that the encoder and decoder are transformer-based architectures, and the decoder's input includes both unmasked tokens and learnable masked tokens to preserve location information.
}
\label{fig:mae_arch}
\vspace{-0.1in}
\end{figure}

MAE~\cite{masked_autoencoder} employs an asymmetric transformer-based encoder-decoder architecture. 
The workflow of MAE, illustrated in Figure~\ref{fig:mae_arch}, involves the following steps:
\begin{itemize}
\itemsep0.3em

    \item The input image is divided into patches, denoted as $x=\{x_i\}_{i=1}^{N}$, where $N$ is the number of patches. The mask is represented as $m=\{m_i\}_{i=1}^{N}$, where each $m_i \in \{0,1\}$. The unmasked region, serving as the actual input to the MAE model, is expressed as $x \odot m = \{m_i \cdot x_i\}_{i=1}^{N}$, where $\odot$ indicates the operation of applying the mask. The masking ratio can be represented as $p = 1 - \frac{\sum_{i=1}^{N} m_i}{N}$.

    \item Only the unmasked patches are flattened and fed into the encoder. The encoder design follows the standard ViT encoder, as depicted in Figure~\ref{fig:vit_arch}. The learnable masked tokens are integrated into the latent representation, alongside the unmasked tokens, and are then fed into the decoder.

    \item Finally, the decoder recovers the masked region and then reconstructs the entire image by substituting the unmasked region with the input patches. The reconstructed image is represented as $x^\prime = x \odot m + mae(x \odot m),$ where $mae(x \odot m)$ is the recovered masked region.


\end{itemize}

\textit{Masking and reconstruction strategy}: Patches have varying importance in the classification process, due to the self-attention mechanism.  In our detection method, we employ a random masking strategy to process the original input. This masking approach enables MAE to capture the information from both the foreground and background, thereby providing a well-reconstructed image. In Section~\ref{masking_strategy}, we selectively mask the foreground or background patches, and demonstrate the superiority of the random masking method. In addition, the MAE model can be employed repeatedly and flexibly to recover different areas of the input. 
This adaptable reconstruction strategy enables ViTGuard to effectively identify adversarial examples from patch attacks, which concentrate perturbations on a minimal number of patches.
The implementation details of the MAE model are explained in Section~\ref{sec:vitguard_config}.

\textit{Visualization of reconstructed images}: Figure~\ref{fig:vis_inputs}(b) in Appendix~\ref{appendix:vis_inputs} visualizes the reconstructed images associated with normal and adversarial inputs. For normal inputs, the reconstructed images introduce imperceptible noise, whereas for adversarial inputs, the reconstructed images exhibit more noticeable noise. 

\subsection{Detector \uppercase\expandafter{\romannumeral1} based on Attention (ViTGuard-I)} \label{detector-1}
While the scaled dot-product attention plays a crucial role in establishing global relationships among different tokens, it has been demonstrated that the raw attention map cannot accurately assess the importance of each token in the classification process~\cite{attn_interpretable}. As a solution, we use a more effective method known as \textit{attention rollout}~\cite{rollout}  
to quantify the information flow through the self-attention mechanism. Attention rollout tracks the information flow by considering residual connections in transformer layers. Given the Value in layer $l-1$, denoted as $V_{l-1}$, 
and the raw attention map $W_{att}(l) \in \mathbb{R}^{(N+1) \times (N+1)}$ in layer $l$, 
the Value in layer $l$ is represented as $V_{l} = (W_{att}(l)+I)V_{l-1}$, where $I$ is an identity matrix. Thus, the normalized attention map, which takes residual connections into account, can be expressed as 
\begin{equation}
    A_l = 0.5W_{att}(l) + 0.5I .
\end{equation}
Furthermore, by recursively multiplying the attention maps from previous layers, the attention by Attention Rollout in layer $l$, denoted as $\tilde{A}_l$, is 
\begin{equation}
    \tilde{A}_l= \begin{cases}A_l \tilde{A}_{l-1} & \text { if } i>0 \\ A_l & \text { if } l=0\end{cases}, 
\end{equation}
where $\tilde{A}_{l-1}$ is the attention by Attention Rollout in layer $l-1$. As only the CLS token is connected to the MLP head and directly used for classification, we extract the attention vector corresponding to the CLS token and use it for our detector design. Given an input $x$, we denote the attention vector from layer $l$ as $attn_l(x)$. In the rest of the paper, `attention' refers to the attention vector corresponding to the CLS token calculated by Attention Rollout.

If the input is adversarial, the attention divergence between the original image $x$ and the reconstructed image $x^\prime$ is expected to be larger. We use Euclidean distance ($L_2$ norm) to measure the attention divergence:
\begin{equation} \label{eq:reconstruction_error}
d_{attn}(x, x^\prime) =\| attn_l(x)-attn_l(x^\prime)) \|_2.
\end{equation}
If $d_{attn}(x, x^\prime)$ is larger than a predefined threshold, the input is classified as adversarial. Otherwise, the input is classified as normal.

\textit{Intermediate Layer Choice}: The intermediate layers utilized by the detector vary according to different datasets. Traditional detection methods for DNN models typically extract features from deep layers, such as the last layer. However, it has been shown that there is no single best layer for analyzing attention maps and latent representations from ViT models~\cite{teaching_matters}. 
Furthermore, it remains an open problem to interpret the attention patterns learned through supervised training for ViTs~\cite{interpretability_1, interpretability_2}. In practice, We select the layer that attains the best AUC score against the FGSM attack based on the validation dataset. This selected layer is subsequently employed to detect against all other adversarial attacks. This selection approach is feasible since FGSM is a well-established adversarial attack algorithm, known for its simplicity and efficiency in generating adversarial examples. Moreover, in Section~\ref{sec:layer_choice}, we investigate the impact of selecting different transformer layers on the detection of various attacks. We demonstrate that, for optimal detection performance, the selection of the intermediate layer needs to vary depending on the specific datasets and attacks.  

\subsection{Detector \uppercase\expandafter{\romannumeral2} based on CLS Representation (ViTGuard-II)} \label{detector-2}

Besides the attention mechanism, another special feature of ViT is that a classification [CLS] token is added to the input sequence made up of image patches, as illustrated in Figure ~\ref{fig:vit_arch}.  At the end of the transformer encoder, only the CLS representation is directed to the MLP head for label prediction, while the representations of other image patches are not utilized. 
It has been demonstrated that, instead of using CLS, the model can achieve a similar classification performance by applying an averaging layer to the image patch representations and then forwarding the averaging layer's output to the MLP head~\cite{pooling}. However, in this paper, we show the significant ability of the CLS token to enhance ViT's robustness by designing a detector based on CLS representations. 

The process for Detector \uppercase\expandafter{\romannumeral2} is the same as that for Detector \uppercase\expandafter{\romannumeral1}, with the only difference of using a different metric. Given the original image $x$, we first create the reconstructed image $x^\prime$. Then, we send both $x$ and $x^\prime$ to the classifier and compare the difference of their CLS representations. The CLS representations of the original image and the reconstructed image from the $l$-th transformer layer are denoted as $cls_l(x)$ and $cls_l(x^\prime)$, respectively. The difference between CLS representations is measured using Euclidean distance:
\begin{equation} 
d_{cls}(x, x^\prime) = \| cls_l(x)- cls_l(x^\prime) \|_2 
\end{equation}
The input is detected as adversarial if $d_{cls}(x, x^\prime)$ exceeds a predefined threshold. In practice, for Detector \uppercase\expandafter{\romannumeral2}, we utilize the same transformer layer that is chosen for Detector \uppercase\expandafter{\romannumeral1}.

\subsection{Joint Detectors}  \label{sec:strategy_joint_detectors}
To further enhance the detection capability, we can integrate the two detectors.
Given the original input $x$ and its reconstructed counterpart  $x^\prime$, if either Detector \uppercase\expandafter{\romannumeral1} or Detector \uppercase\expandafter{\romannumeral2} identifies the input as adversarial, it is classified as adversarial; otherwise, it is classified as normal. 
We observe slight performance variations between the two individual detectors in diverse scenarios involving different attacks and target models. 
Therefore, joint detection is particularly suited for black-box detection scenarios, where the detection system lacks knowledge about the attack types. 



\section{Evaluation} 

In this section, we first explain the experimental setting and then compare our detection methods with existing ones (refer to Sections~\ref{sec:invidual_detector} and~\ref{sec:joint_patch}). Following this, our focus shifts to examining the impact of important elements in ViTGuard and assessing its effectiveness with limited accessible normal images. Finally, we demonstrate ViTGuard's robustness against possible evasion.

\begin{table*}[t]
\centering
\caption{AUC scores of various attack detection methods: VG-I and VG-II refer to ViTGuard-I and ViTGuard-II, respectively. Patch-F and Attn-F correspond to the Patch-Fool and Attention-Fool patch attacks. The remaining attacks are $L_p$ norm attacks.}

\label{tab:roc_scores}
\resizebox{2.09\columnwidth}{!}{
\begin{tabular}{|ccccccccccccccccc|}
\hline
\multicolumn{1}{|c||}{\multirow{2}{*}{Attack}} & \multicolumn{8}{c||}{ViT-16}     & \multicolumn{8}{c|}{ViT-32}    \\ \cline{2-17} 
\multicolumn{1}{|c||}{}                        & \multicolumn{1}{c|}{KD+BU}  & \multicolumn{1}{c|}{LID}    & \multicolumn{1}{c|}{FS}   & \multicolumn{1}{c|}{MSCN} & \multicolumn{1}{c|}{DAE}  & \multicolumn{1}{c|}{GAN} & \multicolumn{1}{c|}{\textbf{VG-\uppercase\expandafter{\romannumeral1}}} & \multicolumn{1}{c||}{\textbf{VG-\uppercase\expandafter{\romannumeral2}}}                           & \multicolumn{1}{c|}{KD+BU}  & \multicolumn{1}{c|}{LID}    & \multicolumn{1}{c|}{FS}    & \multicolumn{1}{c|}{MSCN} & \multicolumn{1}{c|}{DAE} & \multicolumn{1}{c|}{GAN}  & \multicolumn{1}{c|}{\textbf{VG-\uppercase\expandafter{\romannumeral1}}} & \textbf{VG-\uppercase\expandafter{\romannumeral2}}                           \\ \hline
\multicolumn{17}{|c|}{CIFAR-10}    
\\ \hline
\multicolumn{1}{|c||}{FGSM}                    & \multicolumn{1}{c|}{0.7319}      & \multicolumn{1}{c|}{0.9265}    & \multicolumn{1}{c|}{0.5466}  &\multicolumn{1}{c|}{\textbf{1.0000}} & \multicolumn{1}{c|}{0.6470} & \multicolumn{1}{c|}{0.7586} & \multicolumn{1}{c|}{\textbf{1.0000}}       &\multicolumn{1}{c||}{0.9927}        & \multicolumn{1}{c|}{0.6782}      & \multicolumn{1}{c|}{0.8710}    & \multicolumn{1}{c|}{0.8102} & \multicolumn{1}{c|}{\textbf{1.0000}} & \multicolumn{1}{c|}{0.6510} & \multicolumn{1}{c|}{0.7554}  & \multicolumn{1}{c|}{0.9990}       & 0.9897       \\ \hline

\multicolumn{1}{|c||}{PGD}                     & \multicolumn{1}{c|}{0.7311}      & \multicolumn{1}{c|}{0.8289}    & \multicolumn{1}{c|}{0.7552}  &\multicolumn{1}{c|}{0.9763} & \multicolumn{1}{c|}{0.5158} & \multicolumn{1}{c|}{0.9848} & \multicolumn{1}{c|}{\textbf{0.9956}}       & \multicolumn{1}{c||}{0.9844}       & \multicolumn{1}{c|}{0.6760}      & \multicolumn{1}{c|}{0.7821}    & \multicolumn{1}{c|}{0.6592}  & \multicolumn{1}{c|}{0.9672} & \multicolumn{1}{c|}{0.5168} & \multicolumn{1}{c|}{0.9831}  & \multicolumn{1}{c|}{0.9848}       & \textbf{0.9913}      \\ \hline

\multicolumn{1}{|c||}{APGD}                    & \multicolumn{1}{c|}{0.7284}      & \multicolumn{1}{c|}{0.9137}    & \multicolumn{1}{c|}{0.7008} &\multicolumn{1}{c|}{0.9984}  & \multicolumn{1}{c|}{0.5667} & \multicolumn{1}{c|}{0.9797}  & \multicolumn{1}{c|}{\textbf{0.9999}}       & \multicolumn{1}{c||}{0.9926}       & \multicolumn{1}{c|}{0.6719}      & \multicolumn{1}{c|}{0.8552}    & \multicolumn{1}{c|}{0.7050}  & \multicolumn{1}{c|}{0.9956}  & \multicolumn{1}{c|}{0.5706}  & \multicolumn{1}{c|}{0.9770} & \multicolumn{1}{c|}{\textbf{0.9998}}       &0.9961        \\ \hline

\multicolumn{1}{|c||}{CW}                      & \multicolumn{1}{c|}{0.6718}      & \multicolumn{1}{c|}{0.6790}    & \multicolumn{1}{c|}{0.9699}  &\multicolumn{1}{c|}{0.6880}  & \multicolumn{1}{c|}{0.5020}  & \multicolumn{1}{c|}{\textbf{0.9932}} & \multicolumn{1}{c|}{0.9215}       & \multicolumn{1}{c||}{0.9906}       & \multicolumn{1}{c|}{0.6751}      & \multicolumn{1}{c|}{0.7148}    & \multicolumn{1}{c|}{0.9199} & \multicolumn{1}{c|}{0.7410}   & \multicolumn{1}{c|}{0.5040} & \multicolumn{1}{c|}{0.9971} & \multicolumn{1}{c|}{0.9465}       &\textbf{0.9988}        \\ \hline

\multicolumn{1}{|c||}{SGM}              & \multicolumn{1}{c|}{0.8613}      & \multicolumn{1}{c|}{0.7000}    & \multicolumn{1}{c|}{0.6576}  &\multicolumn{1}{c|}{\textbf{0.9980}}  & \multicolumn{1}{c|}{0.5361} & \multicolumn{1}{c|}{0.8320} & \multicolumn{1}{c|}{0.9935}      & \multicolumn{1}{c||}{0.9917}       & \multicolumn{1}{c|}{0.7623}      & \multicolumn{1}{c|}{0.6481}    & \multicolumn{1}{c|}{0.7469} & \multicolumn{1}{c|}{\textbf{0.9917}}   & \multicolumn{1}{c|}{0.5370} & \multicolumn{1}{c|}{0.7959} & \multicolumn{1}{c|}{0.9827}       & 0.9831      \\ \hline

\multicolumn{1}{|c||}{SE}              & \multicolumn{1}{c|}{0.7303}      & \multicolumn{1}{c|}{0.7883}    & \multicolumn{1}{c|}{0.6587} &\multicolumn{1}{c|}{\textbf{1.0000}} & \multicolumn{1}{c|}{0.5564} & \multicolumn{1}{c|}{0.8652} & \multicolumn{1}{c|}{0.9996}     & \multicolumn{1}{c||}{0.9946}       & \multicolumn{1}{c|}{0.8677}      & \multicolumn{1}{c|}{0.6314}    & \multicolumn{1}{c|}{0.7952} & \multicolumn{1}{c|}{0.9810}  & \multicolumn{1}{c|}{0.5603} & \multicolumn{1}{c|}{0.7521} & \multicolumn{1}{c|}{\textbf{0.9972}}       & 0.9967     \\ \hline

\multicolumn{1}{|c||}{TR}              & \multicolumn{1}{c|}{0.7032}      & \multicolumn{1}{c|}{0.6669}    & \multicolumn{1}{c|}{0.7968}  &\multicolumn{1}{c|}{0.9860}  & \multicolumn{1}{c|}{0.5170}  & \multicolumn{1}{c|}{0.9011} & \multicolumn{1}{c|}{\textbf{0.9976}}       & \multicolumn{1}{c||}{0.9872}       & \multicolumn{1}{c|}{0.7036}      & \multicolumn{1}{c|}{0.5888}    & \multicolumn{1}{c|}{0.8136} & \multicolumn{1}{c|}{0.9919}   & \multicolumn{1}{c|}{0.5190} & \multicolumn{1}{c|}{0.7382} & \multicolumn{1}{c|}{\textbf{0.9929}}       & 0.9860     \\ \hline

\multicolumn{1}{|c||}{Patch-F}              & \multicolumn{1}{c|}{0.7308}      & \multicolumn{1}{c|}{0.7422}    & \multicolumn{1}{c|}{0.7960} &\multicolumn{1}{c|}{0.5560}    & \multicolumn{1}{c|}{0.5833}    & \multicolumn{1}{c|}{0.9250} & \multicolumn{1}{c|}{\textbf{0.9997}}  & \multicolumn{1}{c||}{0.9940}       & \multicolumn{1}{c|}{0.7256}      & \multicolumn{1}{c|}{0.6666}    & \multicolumn{1}{c|}{0.8873}  & \multicolumn{1}{c|}{0.5437} & \multicolumn{1}{c|}{0.5821}  & \multicolumn{1}{c|}{0.9422} & \multicolumn{1}{c|}{\textbf{0.9981}}       &0.9976        \\ \hline

\multicolumn{1}{|c||}{Attn-F}              & \multicolumn{1}{c|}{0.7390}      & \multicolumn{1}{c|}{0.7541}    & \multicolumn{1}{c|}{0.8070} &\multicolumn{1}{c|}{0.5646}    & \multicolumn{1}{c|}{0.5836}    & \multicolumn{1}{c|}{0.9251} & \multicolumn{1}{c|}{\textbf{0.9998}}  & \multicolumn{1}{c||}{0.9954}       & \multicolumn{1}{c|}{0.7024}      & \multicolumn{1}{c|}{0.6597}    & \multicolumn{1}{c|}{0.8996}  & \multicolumn{1}{c|}{0.5504} & \multicolumn{1}{c|}{0.5816}  & \multicolumn{1}{c|}{0.9453} & \multicolumn{1}{c|}{0.9970}       &\textbf{0.9974}        \\ \hline

\multicolumn{17}{|c|}{CIFAR-100}   \\ \hline
\multicolumn{1}{|c||}{FGSM}                    & \multicolumn{1}{c|}{0.9389}      & \multicolumn{1}{c|}{0.9105}    & \multicolumn{1}{c|}{0.7249} &\multicolumn{1}{c|}{\textbf{0.9999}}  & \multicolumn{1}{c|}{0.6048} & \multicolumn{1}{c|}{0.9218} & \multicolumn{1}{c|}{\textbf{0.9999}}       &\multicolumn{1}{c||}{0.9933}        & \multicolumn{1}{c|}{0.8888}      & \multicolumn{1}{c|}{0.8513}    & \multicolumn{1}{c|}{0.7206} & \multicolumn{1}{c|}{\textbf{1.0000}}  & \multicolumn{1}{c|}{0.6510} & \multicolumn{1}{c|}{0.9049}  & \multicolumn{1}{c|}{\textbf{1.0000}}       &0.9879         \\ \hline

\multicolumn{1}{|c||}{PGD}                     & \multicolumn{1}{c|}{0.8430}      & \multicolumn{1}{c|}{0.8128}    & \multicolumn{1}{c|}{0.6700} &\multicolumn{1}{c|}{0.9711} & \multicolumn{1}{c|}{0.5085} & \multicolumn{1}{c|}{0.9916} & \multicolumn{1}{c|}{\textbf{0.9920}}       &\multicolumn{1}{c||}{0.9794}        & \multicolumn{1}{c|}{0.8345}      & \multicolumn{1}{c|}{0.7553}    & \multicolumn{1}{c|}{0.6016}  & \multicolumn{1}{c|}{0.9672} & \multicolumn{1}{c|}{0.5168} & \multicolumn{1}{c|}{0.9867}  & \multicolumn{1}{c|}{0.9798}       &\textbf{0.9887}        \\ \hline

\multicolumn{1}{|c||}{APGD}                    & \multicolumn{1}{c|}{0.8681}      & \multicolumn{1}{c|}{0.9034}    & \multicolumn{1}{c|}{0.6974} &\multicolumn{1}{c|}{0.9973} & \multicolumn{1}{c|}{0.5278} & \multicolumn{1}{c|}{0.9924}  & \multicolumn{1}{c|}{\textbf{0.9996}}       &\multicolumn{1}{c||}{0.9901}        & \multicolumn{1}{c|}{0.8349}      & \multicolumn{1}{c|}{0.8379}    & \multicolumn{1}{c|}{0.6369}   & \multicolumn{1}{c|}{0.9956} & \multicolumn{1}{c|}{0.5706} & \multicolumn{1}{c|}{0.9897} & \multicolumn{1}{c|}{\textbf{0.9995}}       &0.9946        \\ \hline

\multicolumn{1}{|c||}{CW}                      & \multicolumn{1}{c|}{0.8510}      & \multicolumn{1}{c|}{0.7437}    & \multicolumn{1}{c|}{0.7840} &\multicolumn{1}{c|}{0.8121} & \multicolumn{1}{c|}{0.5045}  & \multicolumn{1}{c|}{\textbf{0.9990}} & \multicolumn{1}{c|}{0.9694}       &\multicolumn{1}{c||}{0.9956}        & \multicolumn{1}{c|}{0.8068}      & \multicolumn{1}{c|}{0.7027}    & \multicolumn{1}{c|}{0.7473}  & \multicolumn{1}{c|}{0.7410} & \multicolumn{1}{c|}{0.5040} & \multicolumn{1}{c|}{0.9962}  & \multicolumn{1}{c|}{0.9447}       & \textbf{0.9984}       \\ \hline

\multicolumn{1}{|c||}{SGM}              & \multicolumn{1}{c|}{0.8291}      & \multicolumn{1}{c|}{0.6811}    & \multicolumn{1}{c|}{0.7072}  &\multicolumn{1}{c|}{\textbf{0.9945}}  & \multicolumn{1}{c|}{0.5284}  & \multicolumn{1}{c|}{0.9070} & \multicolumn{1}{c|}{0.9939}       & \multicolumn{1}{c||}{0.9801}       & \multicolumn{1}{c|}{0.7423}      & \multicolumn{1}{c|}{0.6359}    & \multicolumn{1}{c|}{0.6672}  & \multicolumn{1}{c|}{\textbf{0.9917}} & \multicolumn{1}{c|}{0.5370}  & \multicolumn{1}{c|}{0.8099}  & \multicolumn{1}{c|}{0.9655}       & 0.9678       \\ \hline

\multicolumn{1}{|c||}{SE}              & \multicolumn{1}{c|}{0.8466}      & \multicolumn{1}{c|}{0.6462}    & \multicolumn{1}{c|}{0.7399}  &\multicolumn{1}{c|}{0.9842} & \multicolumn{1}{c|}{0.5140}  & \multicolumn{1}{c|}{0.8940} & \multicolumn{1}{c|}{\textbf{0.9926}}       & \multicolumn{1}{c||}{0.9788}       & \multicolumn{1}{c|}{0.6824}      & \multicolumn{1}{c|}{0.6161}    & \multicolumn{1}{c|}{0.6559}  & \multicolumn{1}{c|}{\textbf{0.9810}}  & \multicolumn{1}{c|}{0.5603}  & \multicolumn{1}{c|}{0.8128}  & \multicolumn{1}{c|}{0.9722}       & 0.9649     \\ \hline

\multicolumn{1}{|c||}{TR}              & \multicolumn{1}{c|}{0.9319}      & \multicolumn{1}{c|}{0.6836}    & \multicolumn{1}{c|}{0.7834} &\multicolumn{1}{c|}{0.9922}  & \multicolumn{1}{c|}{0.5166} & \multicolumn{1}{c|}{0.9414} & \multicolumn{1}{c|}{\textbf{0.9957}}       & \multicolumn{1}{c||}{0.9909}       & \multicolumn{1}{c|}{0.8490}      & \multicolumn{1}{c|}{0.6284}    & \multicolumn{1}{c|}{0.6906}   & \multicolumn{1}{c|}{\textbf{0.9919}} & \multicolumn{1}{c|}{0.5190}  & \multicolumn{1}{c|}{0.8456} & \multicolumn{1}{c|}{0.9782}       & 0.9822      \\ \hline

\multicolumn{1}{|c||}{Patch-F}              & \multicolumn{1}{c|}{0.8871}      & \multicolumn{1}{c|}{0.7257}    & \multicolumn{1}{c|}{0.8835} &\multicolumn{1}{c|}{0.5613} & \multicolumn{1}{c|}{0.5659} & \multicolumn{1}{c|}{0.9739} & \multicolumn{1}{c|}{\textbf{0.9998}}       &\multicolumn{1}{c||}{0.9904}        & \multicolumn{1}{c|}{0.8646}      & \multicolumn{1}{c|}{0.6671}    & \multicolumn{1}{c|}{0.7169}  & \multicolumn{1}{c|}{0.5437} & \multicolumn{1}{c|}{0.5821} & \multicolumn{1}{c|}{0.9757} & \multicolumn{1}{c|}{\textbf{0.9976}}       & 0.9961       \\ \hline

\multicolumn{1}{|c||}{Attn-F}              & \multicolumn{1}{c|}{0.8937}      & \multicolumn{1}{c|}{0.7235}    & \multicolumn{1}{c|}{0.8821} &\multicolumn{1}{c|}{0.5627} & \multicolumn{1}{c|}{0.5658} & \multicolumn{1}{c|}{0.9736} & \multicolumn{1}{c|}{\textbf{0.9994}}       &\multicolumn{1}{c||}{0.9895}        & \multicolumn{1}{c|}{0.8580}      & \multicolumn{1}{c|}{0.6715}    & \multicolumn{1}{c|}{0.7232}  & \multicolumn{1}{c|}{0.5440} & \multicolumn{1}{c|}{0.5647} & \multicolumn{1}{c|}{0.9732} & \multicolumn{1}{c|}{\textbf{0.9955}}       & 0.9952       \\ \hline

\multicolumn{17}{|c|}{Tiny-ImageNet}   \\ \hline
\multicolumn{1}{|c||}{FGSM}                    & \multicolumn{1}{c|}{0.9202}      & \multicolumn{1}{c|}{0.9194}    & \multicolumn{1}{c|}{0.6395}  &\multicolumn{1}{c|}{0.9993} & \multicolumn{1}{c|}{0.6045} & \multicolumn{1}{c|}{0.9130} & \multicolumn{1}{c|}{0.9995}       &\multicolumn{1}{c||}{\textbf{1.0000}}        & \multicolumn{1}{c|}{0.8851}      & \multicolumn{1}{c|}{0.8902}    & \multicolumn{1}{c|}{0.7273}  & \multicolumn{1}{c|}{0.9982} & \multicolumn{1}{c|}{0.5993} & \multicolumn{1}{c|}{0.9111}  & \multicolumn{1}{c|}{0.9998}       &\textbf{1.0000}        \\ \hline

\multicolumn{1}{|c||}{PGD}                     & \multicolumn{1}{c|}{0.7902}      & \multicolumn{1}{c|}{0.8311}    & \multicolumn{1}{c|}{0.6211}  &\multicolumn{1}{c|}{0.7857} & \multicolumn{1}{c|}{0.5096} & \multicolumn{1}{c|}{0.9853} & \multicolumn{1}{c|}{0.9954}       &\multicolumn{1}{c||}{\textbf{0.9999}}        & \multicolumn{1}{c|}{0.7434}      & \multicolumn{1}{c|}{0.7468}    & \multicolumn{1}{c|}{0.5475}  & \multicolumn{1}{c|}{0.7760}  & \multicolumn{1}{c|}{0.5069} & \multicolumn{1}{c|}{0.9802} & \multicolumn{1}{c|}{0.9915}       &\textbf{0.9998}        \\ \hline

\multicolumn{1}{|c||}{APGD}                    & \multicolumn{1}{c|}{0.7621}      & \multicolumn{1}{c|}{0.9086}    & \multicolumn{1}{c|}{0.5919} &\multicolumn{1}{c|}{0.9743} & \multicolumn{1}{c|}{0.5299} & \multicolumn{1}{c|}{0.9873} & \multicolumn{1}{c|}{0.9996}       &\multicolumn{1}{c||}{\textbf{1.0000}}        & \multicolumn{1}{c|}{0.7444}      & \multicolumn{1}{c|}{0.8544}    & \multicolumn{1}{c|}{0.5452}  & \multicolumn{1}{c|}{0.9681}   & \multicolumn{1}{c|}{0.5273} & \multicolumn{1}{c|}{0.9879} & \multicolumn{1}{c|}{\textbf{0.9999}}       &\textbf{0.9999}        \\ \hline

\multicolumn{1}{|c||}{CW}                      & \multicolumn{1}{c|}{0.6272}      & \multicolumn{1}{c|}{0.6826}    & \multicolumn{1}{c|}{0.8576} &\multicolumn{1}{c|}{0.5775}  & \multicolumn{1}{c|}{0.5014} & \multicolumn{1}{c|}{0.9919}  & \multicolumn{1}{c|}{0.9421}       &\multicolumn{1}{c||}{\textbf{0.9892}}        & \multicolumn{1}{c|}{0.7009}      & \multicolumn{1}{c|}{0.6310}    & \multicolumn{1}{c|}{0.8039}  & \multicolumn{1}{c|}{0.5640}  & \multicolumn{1}{c|}{0.5010}  & \multicolumn{1}{c|}{0.9905} & \multicolumn{1}{c|}{0.8539}       & \textbf{0.9911}       \\ \hline

\multicolumn{1}{|c||}{SGM}              & \multicolumn{1}{c|}{0.7999}      & \multicolumn{1}{c|}{0.7247}    & \multicolumn{1}{c|}{0.6584}  &\multicolumn{1}{c|}{0.8288}  & \multicolumn{1}{c|}{0.5208} & \multicolumn{1}{c|}{0.9058} & \multicolumn{1}{c|}{0.9085}       & \multicolumn{1}{c||}{\textbf{0.9627}}       & \multicolumn{1}{c|}{0.7675}      & \multicolumn{1}{c|}{0.6313}    & \multicolumn{1}{c|}{0.6923} & \multicolumn{1}{c|}{0.8323}  & \multicolumn{1}{c|}{0.5212} & \multicolumn{1}{c|}{0.8636}  & \multicolumn{1}{c|}{\textbf{0.9022}}       & 0.7548       \\ \hline

 \multicolumn{1}{|c||}{SE}              & \multicolumn{1}{c|}{0.8410}      & \multicolumn{1}{c|}{0.6569}    & \multicolumn{1}{c|}{0.6692} &\multicolumn{1}{c|}{0.7817} & \multicolumn{1}{c|}{0.5083} & \multicolumn{1}{c|}{0.9302}  & \multicolumn{1}{c|}{0.9518}       & \multicolumn{1}{c||}{\textbf{0.9902}}       & \multicolumn{1}{c|}{0.6883}      & \multicolumn{1}{c|}{0.5738}    & \multicolumn{1}{c|}{0.6387} & \multicolumn{1}{c|}{0.7981}  & \multicolumn{1}{c|}{0.5089} & \multicolumn{1}{c|}{0.8345}  & \multicolumn{1}{c|}{\textbf{0.9564}}       & 0.8887      \\ \hline

\multicolumn{1}{|c||}{TR}              & \multicolumn{1}{c|}{0.8793}      & \multicolumn{1}{c|}{0.7145}    & \multicolumn{1}{c|}{0.7074}  &\multicolumn{1}{c|}{0.8226}  & \multicolumn{1}{c|}{0.5087} & \multicolumn{1}{c|}{0.9682} & \multicolumn{1}{c|}{0.9634}       & \multicolumn{1}{c||}{\textbf{0.9934}}       & \multicolumn{1}{c|}{0.7543}      & \multicolumn{1}{c|}{0.6077}    & \multicolumn{1}{c|}{0.6917} & \multicolumn{1}{c|}{0.8251}  & \multicolumn{1}{c|}{0.5094} & \multicolumn{1}{c|}{0.9036}  & \multicolumn{1}{c|}{\textbf{0.9657}}       & 0.9220      \\ \hline

\multicolumn{1}{|c||}{Patch-F}              & \multicolumn{1}{c|}{0.7615}      & \multicolumn{1}{c|}{0.6856}    & \multicolumn{1}{c|}{0.8157} &\multicolumn{1}{c|}{0.5292}   & \multicolumn{1}{c|}{0.5452} & \multicolumn{1}{c|}{0.9710} & \multicolumn{1}{c|}{\textbf{0.9941}}       &\multicolumn{1}{c||}{0.7795}        & \multicolumn{1}{c|}{0.7394}      & \multicolumn{1}{c|}{0.6232}    & \multicolumn{1}{c|}{0.7450} & \multicolumn{1}{c|}{0.5284}  & \multicolumn{1}{c|}{0.5489} & \multicolumn{1}{c|}{\textbf{0.9670}}  & \multicolumn{1}{c|}{0.9307}       &0.8846       \\ \hline

\multicolumn{1}{|c||}{Attn-F}              & \multicolumn{1}{c|}{0.7723}      & \multicolumn{1}{c|}{0.6948}    & \multicolumn{1}{c|}{0.8135} &\multicolumn{1}{c|}{0.5306}   & \multicolumn{1}{c|}{0.5433} & \multicolumn{1}{c|}{0.9679} & \multicolumn{1}{c|}{\textbf{0.9920}}       &\multicolumn{1}{c||}{0.7574}        & \multicolumn{1}{c|}{0.7240}      & \multicolumn{1}{c|}{0.6204}    & \multicolumn{1}{c|}{0.7644} & \multicolumn{1}{c|}{0.5251}  & \multicolumn{1}{c|}{0.5464} & \multicolumn{1}{c|}{\textbf{0.9688}}  & \multicolumn{1}{c|}{0.9246}       &0.8773       \\ \hline
\end{tabular}
}
\end{table*}

\subsection{Experimental Setup}
\subsubsection{Datasets and Experimental Environment}
We utilize three popular image datasets with varying image dimensions and numbers of classes. \textbf{CIFAR-10~\cite{cifar10}} comprises ten object classes, with each image having a size of  $32 \times 32$. \textbf{CIFAR-100~\cite{cifar10}} includes 100 object classes and shares the same image size as CIFAR-10. \textbf{Tiny-ImageNet~\cite{tiny-imagenet}} contains color images labeled across 200 classes, with each image having a size of $64 \times 64$. More details about the datasets are provided in Appendix~\ref{appendix_dataset}.

\subsubsection{ViTGuard Configurations} \label{sec:vitguard_config}
\textbf{MAE}: We use the ViT-MAE-Base model with a patch size of $16 \times 16$ for input reconstruction. The encoder is composed of 12 transformer blocks and 12 attention heads, while the decoder consists of 8 transformer blocks and 16 attention heads. MAE models are exclusively trained using clean training data samples. The default masking strategy is random masking, with a default masking ratio of 0.5. The impact of different masking strategies and ratios on detection performance will be presented in Sections~\ref{masking_strategy} and~\ref{masking_ratio}.
MAE models are trained from scratch for 250 epochs on CIFAR-10 and CIFAR-100, and for 500 epochs on Tiny-ImageNet, using the Adam optimizer. The learning rate, set at $1.5\times10^{-4}$, gradually decreases based on cosine annealing. For patch attacks, we apply the MAE model twice to recover the entire image. This is necessitated by the attack's perturbation of a minimal number of patches – four and one patches, respectively, for the ViT-16 and ViT-32 target models, considering the total patch number of 196 and 49. 
In preliminary experiments with $L_p$ norm attacks on CIFAR-100 (refer to Appendix~\ref{sec:pre_half_entire}), we observed that full input recovery—applying the MAE model twice—marginally improved AUC scores, ranging from 0.0001 to 0.007, compared to half input recovery. Considering this minimal improvement and the increased computational overhead, we opt to recover only half of the input in evaluations for $L_p$ norm attacks. Importantly, this approach aims to balance detection effectiveness with efficiency, but it does not imply that our detection method depends on the attack type.
\textbf{Our detectors}: As described in Section~\ref{detector-1}, for each dataset, we select the transformer layer with the highest AUC score for ViTGuard-I under FGSM. The chosen layer is then utilized for feature extraction in both ViTGuard-I and ViTGuard-II to detect various adversarial attacks. 
In the evaluation, we use the last transformer layer for CIFAR-10 and CIFAR-100, and the second layer for Tiny-ImageNet. In Section~\ref{sec:layer_choice}, we will illustrate that no single layer is optimal across all adversarial attacks, especially when our detectors have no prior knowledge of the specific attacks.

\subsubsection{Evaluation Metrics}
The detection mechanism works as a binary classifier, and thus the detection accuracy \footnote{The detection is evaluated based on successful adversarial samples, where the original input is correctly classified but the corresponding adversarial sample is misclassified.} is evaluated using the following metrics.
\textbf{True Positive Rate (TPR)} represents the percentage of the adversarial samples that are correctly detected as adversarial. 
\textbf{False Positive Rate (FPR)} represents the percentage of the normal samples that are misclassified as adversarial. 
\textbf{Receiver Operator Characteristic (ROC) Curve}  is used to evaluate the detection capability by plotting TPR against FPR while varying the threshold.
\textbf{Area Under the ROC Curve (AUC) Score} measures the area under the ROC Curve and ranges from 0 to 1. A higher AUC score corresponds to a higher TPR at a fixed FPR, indicating better detection performance.


\subsubsection{Settings of Baseline Detection}
We compare ViTGuard with seven existing detection methods: ARMRO~\cite{armro}, KD+BU~\cite{KDBU}, LID~\cite{Lid}, FS~\cite{Feature_squeezing}, MSCN~\cite{MSCN}, DAE~\cite{magnet}, and GAN~\cite{defense-gan}. ARMRO, state-of-the-art for defending against patch attacks on ViTs, detects and masks adversarial patches without altering the ViT model. Notably, ARMRO is specifically tailored for patch attacks and is not applicable to $L_p$ norm attacks. Thus, we compare ARMRO with ViTGuard only for patch attacks, excluding comparisons with other detection methods for $L_p$ norm attacks.
Except for ARMRO, the other methods were originally designed for $L_p$ norm attacks. FS, DAE, and GAN are unsupervised detection methods, while KD+BU, LID, and MSCN are supervised methods. Although KD+BU and LID have been shown to be ineffective against stronger attacks~\cite{CW} in CNN models, we evaluate their performance in protecting ViTs as they are well-established baselines in adversarial attack detection. DAE and GAN are used as image recovery techniques. An adversarial input is identified if the difference between the original and recovered images exceeds a predefined threshold. For DAE, we use a reconstruction error-based detector, and for GAN, we use a probability divergence-based detector with a softmax temperature of 10. Notably, there is a lack of specifically designed detection mechanisms for $L_p$ norm attacks in ViT models, so we use detection methods originally designed for CNN models. Please refer to Appendix~\ref{appendix_existing} for more implementation details.

\subsection{Comparison with Existing $L_p$ Norm Attack Detection} \label{sec:invidual_detector}

In this section, we compare ViTGuard with detection methods that were designed for $L_p$ norm attacks, and we further test their ability to detect patch attacks.
Table~\ref{tab:roc_scores} summarizes the detection performance using AUC scores, which reflect their overall detection capability at varying thresholds. 
To visually compare the TPRs, Figures~\ref{fig:roc_curves_cifar} and~\ref{fig:roc_curves_tiny} in Appendix~\ref{appendix_roc} show ROC curves with a constrained FPR of 0.2 for various attacks. Moreover, ViTGuard-I and ViTGuard-II exhibit a slight difference in detection performance, as shown in Table~\ref{tab:roc_scores}, due to their utilization of different features. To further enhance detection capability, we combined both detectors, with the results presented in Appendix~\ref{appendix_tpr_fpr}.

\textit{Performance on $L_p$ norm attacks.}
The AUC scores and ROC curves demonstrate that ViTGuard-I and ViTGuard-II consistently exhibit excellent detection performance, with minor discrepancies between them under different attack scenarios. In contrast, existing detection methods either perform poorly across all attacks against ViT models or excel in specific attacks while underperforming in others. For example, the DAE method has limited image recovery capability, which introduces significant noise in normal inputs within a large feature space, leading to inaccurate detection; hence, it is excluded from further analysis. GAN shows a stronger capability to purify adversarial perturbations but still underperforms on certain attacks, such as black-box attacks.
For $L_p$ norm attacks, ViTGuard improves AUC scores compared to existing methods, with increments of up to 0.0789, 0.1916, and 0.2238 for CIFAR-10, CIFAR-100, and Tiny-ImageNet, respectively.
Furthermore, it is observed that the CLS-based detector exhibits superior performance on datasets with larger feature spaces (i.e., larger original image sizes before resizing) due to the presence of more complex attention correlations. Conversely, the attention-based detector is more effective in smaller feature spaces, where simpler correlations facilitate more accurate decisions. To reduce computational demands for $L_p$ norm attacks in practice, the attention-based detector can be prioritized for smaller feature spaces, while the CLS-based detector can be prioritized for larger ones.

\textit{Performance on patch attacks.} Similarly, for Patch-Fool and Attention-Fool attacks, ViTGuard detectors demonstrate superior performance, achieving AUC scores greater than 0.99 in most cases. Specifically, the attention-based detector consistently outperforms the CLS-based detector in the context of patch attacks, as these attacks are inherently designed to manipulate the attention mechanism.
In contrast, existing detection methods are largely ineffective against patch attacks. Among them, GAN shows relatively fair performance, with AUC scores ranging from 0.9250 to 0.9757.

\textbf{Superiority of ViTGuard detectors:}
Although an effective image reconstruction model is critical, detectors also need to consider the unique structural characteristics of the models. ViTGuard detectors utilize intrinsic features of ViT models. In contrast, previous work~\cite{HGD, magnet, Feature_squeezing} generally relies on loss of reconstruction and probability divergence, neglecting the specific features of the target models. To exclusively demonstrate the superiority of ViTGuard detectors, we use the same MAE model for image reconstruction but employ different metrics-based detectors. In this way, the only distinction lies in the design of their respective detectors. Table~\ref{tab:metrics_comparison} compares different detectors using the ViT-16 model with Tiny-ImageNet. We observe that ViTGuard detectors outperform other detectors across various attacks, despite utilizing the same MAE model for image reconstruction. ViTGuard detectors show improvements in AUC scores compared to other detectors with the highest performance, with relative increases of 11. 2\%, 4. 6\%, 20. 0\%, 9. 9\%, and 7. 3\% for PGD, CW, Patch-Fool, SGM, and TR attacks, respectively.

\begin{table}[t]
\caption{AUC scores for detectors based on different metrics using the same MAE model for image reconstruction. RL, $\text{PD}_{T10}$ and $\text{PD}_{T40}$ represent detectors based on reconstruction loss and probability divergence with Softmax temperatures of 10 and 40, respectively.}
\label{tab:metrics_comparison}
\resizebox{\columnwidth}{!}{
\begin{tabular}{|c|c|c|c|c|c|}
\hline
Attack & ViTGuard-I & ViTGuard-II & RL & $\text{PD}_{T10}$ & $\text{PD}_{T40}$ \\ \hline
PGD    & 0.9954     & \textbf{0.9999}      & 0.8994      &0.8895                & 0.8969               \\ \hline
CW     &0.9421     &\textbf{0.9892}      &0.6267      &0.9460                &0.9349                \\ \hline
Patch-Fool    &\textbf{0.9941}     &0.7795      &0.6678      &0.8282                &0.8281                \\ \hline
SGM    &0.9085     &\textbf{0.9627}      &0.8763      &0.7242                &0.7361                \\ \hline
TR    &0.9634     &\textbf{0.9934}      &0.9254      &0.8432                &0.8608                \\ \hline
\end{tabular} }
\vspace{-0.1in}
\end{table}

\textbf{Takeaway:} 
ViTGuard demonstrates robust performance across various attacks, though it exhibits slightly lower performance than other detection methods in some specific cases. Moreover, existing detection methods designed for $L_p$ norm attacks are ineffective for detecting patch attacks. The effectiveness of ViTGuard's detection is attributed not only to its powerful and flexible image reconstruction strategy, which effectively mitigates adversarial perturbations, but also to its use of attention-based and CLS-based detectors.

\subsection{Comparison with Existing Patch Attack Detection} \label{sec:joint_patch}
\begin{figure}[t]
\centerline{\includegraphics[width=\linewidth]{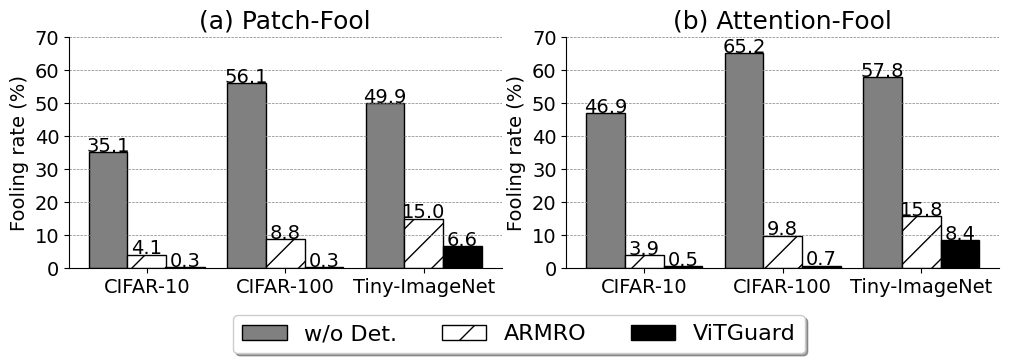}}
\caption{Performance comparison between ViTGaurd and ARMRO under patch attacks against ViT-16 models. `w/o Det.' represents attacks' fooling rates without any detection.}
\label{fig:det_patch}
\vspace{-0.1in}
\end{figure}

To further evaluate ViTGuard's efficacy in patch attacks, we compare it with ARMRO~\cite{armro}, a defense explicitly targeting such attacks against ViT models. ARMRO works by detecting adversarial patches and masking them with the input average. Since ARMRO does not directly detect adversarial examples, we utilize a metric named \textbf{fooling rate} to ensure a fair comparison. The fooling rate for ARMRO measures the percentage of adversarial examples correctly classified out of the total, whereas for ViTGuard, it is the percentage of adversarial examples that evade the joint detectors. Figure~\ref{fig:det_patch} compares the fooling rates of ViTGuard and ARMRO under Patch-Fool and Attention-Fool attacks. The threshold for ViTGuard joint detectors is set to limit FPR to 0.01. It is evident that ViTGuard outperforms ARMRO in detecting patch attacks. For instance, for CIFAR-10 and CIFAR-100, the fooling rates of ViTGuard are below 1\%, while the fooling rates of ARMRO range from 3.9\% to 9.8\%.

\textbf{Takeaway:} ViTGuard is effective in detecting patch attacks, with perturbations confined to only 2\% of the input patches.

\subsection{Impact of key elements in ViTGuard}

\subsubsection{Intermediate Layer Choice} \label{sec:layer_choice}
\begin{figure}[t]
\centerline{\includegraphics[width=\linewidth]{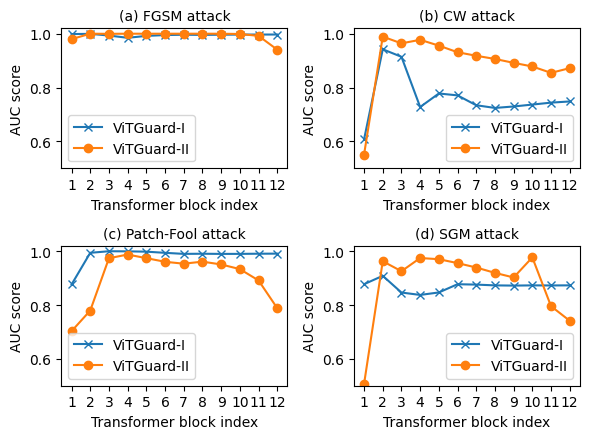}}
\caption{The impact of choosing different transformer blocks on AUC scores for the ViT-16 model with the Tiny-ImageNet dataset. }
\label{fig:layer_selection}
\end{figure}

Rather than fixing a default transformer layer, here we explore the impact of intermediate layer selection on the detection capability under various attacks, as illustrated in Figure~\ref{fig:layer_selection}. We assess AUC scores under four types of adversarial attacks, including FGSM (gradient-based), CW (optimization-based), Patch-Fool (attention-aware), and SGM (transfer-based). 

For the FGSM attack, the AUC scores of both detectors, constructed using each layer, are similar. In the case of the CW attack, both detectors generally demonstrate better performance with shallow layers, reaching the highest AUC score with the second layer. 
For the Patch-Fool attack, ViTGuard-\uppercase\expandafter{\romannumeral1} consistently outperforms ViTGuard-\uppercase\expandafter{\romannumeral2}, achieving AUC scores close to 1 for all layers except the first layer.
 ViTGuard-\uppercase\expandafter{\romannumeral1}, relying on attention, proves to be more effective in detecting the Patch-Fool attack, since the Patch-Fool attack is designed to disrupt the attention maps of images.
For the SGM attack, ViTGuard-\uppercase\expandafter{\romannumeral1} achieves higher performance using features extracted from the second to the tenth layers.
 Overall, no single layer can constantly achieve the best detection performance under different attacks. Therefore, when evaluating our detectors in Section~\ref{sec:invidual_detector}, we do not seek the optimal detection performance, considering that AUC scores are already close to 1 with the default layer selected based on FGSM.

\subsubsection{Impact of Masking Strategies} \label{masking_strategy}
\begin{figure}[t]
\centerline{\includegraphics[width=\linewidth]{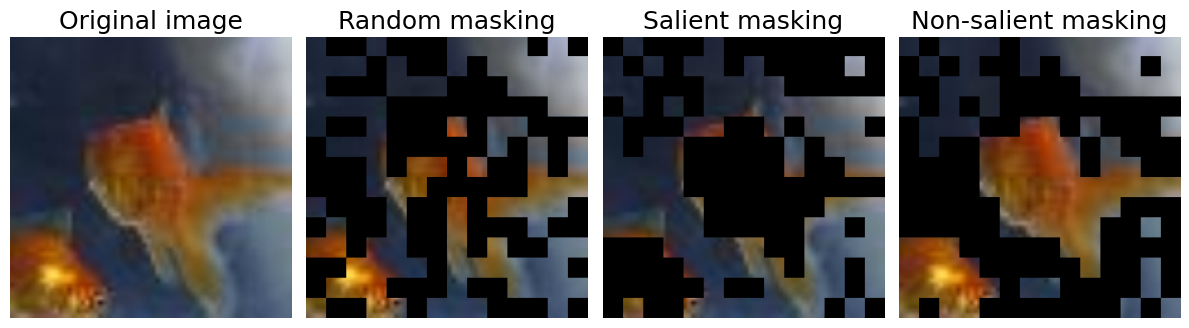}}
\caption{Masked images with different masking strategies.}
\label{fig:masking_strategies}
\end{figure}

Here we investigate the impact of different masking strategies, including salient (foreground) masking and non-salient (background) masking, on the detection performance.  (1) \textit{Salient masking}: The foreground area, which contains important information about the object, is masked. Only the background patches are forwarded to the MAE model. (2) \textit{Non-salient masking}: The background area is masked, and only the foreground patches are sent to the MAE model.
We use the self-supervised ViT model, DINO (self-\textbf{di}stillation with \textbf{no} labels)~\cite{DINO}, to create salient and non-salient masks, since it has been demonstrated that DINO can accurately segment objects from the background. Figure~\ref{fig:masking_strategies} shows the masked images with different masking strategies.

Table~\ref{tab:auc_masking} lists the AUC scores of the detection methods using different masking strategies under $L_p$ norm attacks.  For $L_p$ norm attacks, masking strategies directly affect the image area where adversarial perturbations are purified.
The random masking strategy always achieves the highest AUC scores, with the non-salient masking strategy following closely as the second highest, and the salient masking strategy yields the lowest AUC scores.  These outcomes are reasonable, since random masking retains information from both the object and background. In contrast, other masking strategies restrict input information to either the object or the background, thereby imposing limitations on the effective recovery of masked patches. 

\begin{table}[t]
\caption{AUC scores for different masking strategies on ViT-16 with Tiny-ImageNet: non-salient, salient, and random masking strategies are denoted as Non-Sal, Sal, and Rand, respectively.}
\label{tab:auc_masking}
\resizebox{\columnwidth}{!}{
\begin{tabular}{|c|ccc|ccc|}
\hline
           & \multicolumn{3}{c|}{ViTGuard-\uppercase\expandafter{\romannumeral1}}                                     & \multicolumn{3}{c|}{ViTGuard-\uppercase\expandafter{\romannumeral2}}                                 \\ \hline
Attack     & \multicolumn{1}{c|}{Non-Sal} & \multicolumn{1}{c|}{Sal}    & Rand   & \multicolumn{1}{c|}{Non-Sal} & \multicolumn{1}{c|}{Sal} & Rand   \\ \hline
FGSM       & \multicolumn{1}{c|}{0.9935}  & \multicolumn{1}{c|}{0.9240} & \textbf{0.9995} & \multicolumn{1}{c|}{0.9998}        & \multicolumn{1}{c|}{0.9925}    & \textbf{1.0000} \\ \hline
PGD        & \multicolumn{1}{c|}{0.9741}  & \multicolumn{1}{c|}{0.8546} & \textbf{0.9954} & \multicolumn{1}{c|}{0.9950}        & \multicolumn{1}{c|}{0.9576}    & \textbf{0.9999} \\ \hline
APGD       & \multicolumn{1}{c|}{0.9946}  & \multicolumn{1}{c|}{0.9378} & \textbf{0.9996} & \multicolumn{1}{c|}{0.9998}        & \multicolumn{1}{c|}{0.9890}    & \textbf{1.0000} \\ \hline
CW         & \multicolumn{1}{c|}{0.8785}  & \multicolumn{1}{c|}{0.7456} & \textbf{0.9421} & \multicolumn{1}{c|}{0.9386}        & \multicolumn{1}{c|}{0.8800}    & \textbf{0.9892}\\ \hline
SGM & \multicolumn{1}{c|}{0.7659}        & \multicolumn{1}{c|}{0.7077}     & \textbf{0.9085} & \multicolumn{1}{c|}{0.8595}     & \multicolumn{1}{c|}{0.8435}   & \textbf{0.9627} \\ \hline
SE & \multicolumn{1}{c|}{0.8059}        & \multicolumn{1}{c|}{0.7626}     & \textbf{0.9518} & \multicolumn{1}{c|}{0.9234}     & \multicolumn{1}{c|}{0.9225}   & \textbf{0.9902} \\ \hline
TR & \multicolumn{1}{c|}{0.8326}        & \multicolumn{1}{c|}{0.7883}     & \textbf{0.9634} & \multicolumn{1}{c|}{0.9359}     & \multicolumn{1}{c|}{0.9352}   & \textbf{0.9934} \\ \hline
\end{tabular} }
\end{table}

\subsubsection{Impact of Masking Ratios} \label{masking_ratio}
The masking ratio determines the proportion of input data for training an MAE model for reconstruction. A lower masking ratio allows the MAE model to recover masked patches more effectively by retaining more information as input. However, it could be difficult to discriminate between adversarial and clean samples, given a small number of recovered patches following the normal distribution.
 On the other hand, a higher masking ratio can mitigate the adversarial effect to a greater extent, but it introduces additional reconstruction loss into the reconstructed image.
 Therefore, we investigate the impact of different masking ratios on the detection performance against $L_p$ norm attacks using the ViT-16 model with Tiny-ImageNet. 
 
 Table~\ref{tab:auc_masking_ratio} presents the AUC scores for masking ratios of 0.25, 0.5, and 0.75. In most cases, the detection method with a masking ratio of 0.5 always achieves the best AUC scores. The masking ratios of 0.25 and 0.75 
 yield similar but relatively lower AUC scores. These results suggest that masking half of the patches can effectively mitigate the adversarial effect without introducing excessive reconstruction loss. However, the masking ratio determines the amount of input data used to train the MAE model. When prioritizing computational efficiency, a higher masking ratio would be a better choice, given the negligible difference in detection performance in most cases. 

\subsection{Generalizability on ViT Variants}
Multiple variants~\cite{deit,cait,swin,T2T-ViT,PVT} of the ViT architecture have been developed to increase classification accuracy, improve data efficiency, and enhance model robustness. This study explores the generalizability of ViTGuard across different ViT variants by adapting attention-based and CLS-based detectors to two prominent ViT models: DeiT~\cite{deit} and CaiT~\cite{cait}.
DeiT incorporates a distillation token alongside the CLS token, enabling learning from a teacher model and achieving strong performance with smaller datasets.
CaiT differentiates between self-attention and class-attention layers by first employing self-attention without a CLS token. The CLS token is then introduced in the later layers to effectively utilize relevant information for classification.
The details on the implementation of the DeiT and CaiT models, along with their classification accuracies under various attacks, are given in Appendix~\ref{appendix_variants}.

Table~\ref{tab:auc_variants} presents the AUC scores of the two detectors across four representative attacks: PGD (gradient-based), CW (optimization-based), SGM (transfer-based), and Patch-Fool (patch attack). The results indicate that the CLS-based detector (ViTGuard-II) demonstrates superior generalizibility on DeiT and CaiT compared to the attention-based detector (ViTGuard-I).
This can be attributed to the architectural modifications introduced in the ViT variants, which influence the attention map and potentially reduce the effectiveness of attention-based detectors. Conversely, ViTGuard-II relies on the CLS token, which encapsulates the most critical information related to classification, corresponding to the primary objective of adversarial attacks to induce misclassification. Therefore, the CLS-based detector is better suited to maintain robust performance despite architectural variations in ViT models.


\begin{table}[t]
\caption{AUC scores for different masking ratios on ViT-16 with Tiny-ImageNet.}
\label{tab:auc_masking_ratio}
\resizebox{\columnwidth}{!}{
\begin{tabular}{|c|ccc|ccc|}
\hline
\multirow{3}{*}{Attack} & \multicolumn{3}{c|}{ViTGuard-\uppercase\expandafter{\romannumeral1}}                                    & \multicolumn{3}{c|}{ViTGuard-\uppercase\expandafter{\romannumeral2}}                                   \\ \cline{2-7} 
                        & \multicolumn{3}{c|}{Masking ratio}                                 & \multicolumn{3}{c|}{Masking ratio}                                 \\ \cline{2-7} 
                        & \multicolumn{1}{c|}{0.25}   & \multicolumn{1}{c|}{0.50}   & 0.75   & \multicolumn{1}{c|}{0.25}   & \multicolumn{1}{c|}{0.50}   & 0.75   \\ \hline
FGSM                    & \multicolumn{1}{c|}{0.9950} & \multicolumn{1}{c|}{\textbf{0.9995}} & 0.9971 & \multicolumn{1}{c|}{\textbf{1.0000}} & \multicolumn{1}{c|}{\textbf{1.0000}} & \textbf{1.0000} \\ \hline
PGD                     & \multicolumn{1}{c|}{0.9801} & \multicolumn{1}{c|}{\textbf{0.9954}} & 0.9868 & \multicolumn{1}{c|}{\textbf{0.9999}} & \multicolumn{1}{c|}{\textbf{0.9999}} & 0.9998 \\ \hline
APGD                    & \multicolumn{1}{c|}{0.9967} & \multicolumn{1}{c|}{\textbf{0.9996}} & 0.9983 & \multicolumn{1}{c|}{\textbf{1.0000}} & \multicolumn{1}{c|}{\textbf{1.0000}} & \textbf{1.0000} \\ \hline
CW                      & \multicolumn{1}{c|}{0.9379} & \multicolumn{1}{c|}{\textbf{0.9421}} & 0.9333 & \multicolumn{1}{c|}{\textbf{0.9906}} & \multicolumn{1}{c|}{0.9892} & 0.9850 \\ \hline
SGM              & \multicolumn{1}{c|}{\textbf{0.9101}} & \multicolumn{1}{c|}{0.9085} & 0.8209 & \multicolumn{1}{c|}{0.9566} & \multicolumn{1}{c|}{\textbf{0.9627}} & 0.9610 \\ \hline
SE              & \multicolumn{1}{c|}{0.9517} & \multicolumn{1}{c|}{\textbf{0.9518}} & 0.8470 & \multicolumn{1}{c|}{0.9886} & \multicolumn{1}{c|}{\textbf{0.9902}} & 0.9883 \\ \hline
TR              & \multicolumn{1}{c|}{0.9633} & \multicolumn{1}{c|}{\textbf{0.9634}} & 0.8756 & \multicolumn{1}{c|}{0.9922} & \multicolumn{1}{c|}{\textbf{0.9934}} & 0.9920 \\ \hline
\end{tabular}
}
\end{table}

\begin{table}[]
\caption{AUC scores of ViTGuard detectors applied to ViT variants on the Tiny-ImageNet dataset.}
\label{tab:auc_variants}
\begin{tabular}{|c|cc|cc|}
\hline
\multirow{2}{*}{Attack} & \multicolumn{2}{c|}{DeiT}                     & \multicolumn{2}{c|}{CaiT}                     \\ \cline{2-5} 
                        & \multicolumn{1}{c|}{VG-I} & VG-II & \multicolumn{1}{c|}{VG-I} & VG-II \\ \hline
                        
PGD & \multicolumn{1}{c|}{0.9728}  &  
\textbf{1.0000}           
& \multicolumn{1}{c|}{0.7899}           &\textbf{0.9996}             \\ \hline
CW  & \multicolumn{1}{c|}{0.8723}   &\textbf{0.9993}             & \multicolumn{1}{c|}{0.7230}           & \textbf{1.0000}            \\ \hline
SGM         & \multicolumn{1}{c|}{0.8449}   &\textbf{0.9901}             & \multicolumn{1}{c|}{0.7093}           & \textbf{0.9806}            \\ \hline
Patch-Fool              & \multicolumn{1}{c|}{0.8961}           & \textbf{0.9981}           & \multicolumn{1}{c|}{0.6873}           & \textbf{0.9010}            \\ \hline
\end{tabular} 
\end{table}

\subsection{Pre-trained Masked Autoencoder}
\begin{figure}[t]
\vspace{-0.1in}
\centerline{\includegraphics[width=\linewidth]{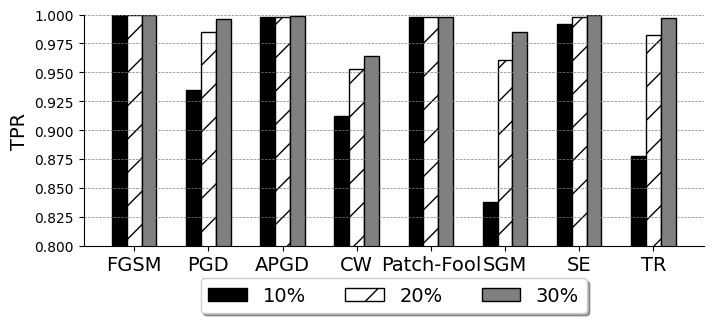}}
\caption{TPRs of joint detectors across different percentages of available training data, ranging from 10\% to 30\%.}
\label{fig:tpr_pretrained}
\vspace{-0.1in}
\end{figure}

While existing unsupervised detection methods~\cite{magnet, HGD, defense-gan, nic} typically utilize the entire clean dataset to train their detectors, we explore a scenario in which ViTGuard detectors have access to only a small fraction of the clean data for a specific task. In such cases, we leverage a pre-trained MAE model, fine-tuning it with the available training data. 
We employ the pre-trained MAE model from Tiny-ImageNet~\footnote{The MAE models trained on different datasets will be publicly available upon the publication of this paper. } and fine-tune it for the CIFAR-10 task with the ViT-16 model. The MAE model is fine-tuned for 250 epochs with a learning rate of $1.5\times10^{-4}$ decreasing based on cosine annealing. Figure~\ref{fig:tpr_pretrained} shows the TPRs of joint detectors under an FPR of 0.05, considering different percentages of available training data ranging from 10\% to 30\%. 
In Figure~\ref{fig:tpr_pretrained}, it is evident that TPRs can approach or reach 1 with only 10\% of data used for training the MAE model under FGSM, APGD, and Patch-Fool attacks. For the remaining attacks, TPRs gradually increase with more available training data. Therefore, the pre-trained masked autoencoder serves as a strong foundation for ViTGuard detectors, making them effective even with a limited amount of clean data samples.


\subsection{Robustness against Evasion}
\label{sec:adaptive_attacks}
In this section, we assume that attackers have full knowledge of our proposed detection methods, including the architecture and parameters of the MAE model, and the victim classifier, as well as the design principles of ViTGuard-I and ViTGuard-II. 
Inspired by adaptive attacks in~\cite{nic, Lid}, we employ the strongest white-box attack, CW, and design a CW-based adaptive attack to evade ViTGuard detectors.
In our adaptive attack approach, we design a new adversarial loss function that accounts for the distances in attention and CLS representations from each transformer block between adversarial and reconstructed images. The detailed loss function is presented in Appendix~\ref{sec:appendix_adaptive}.


To determine the optimal coefficients for the attention term ($\beta_{attn}$) and the CLS term ($\beta_{cls}$), we explore the range $[10^{-5}, 10^3]$ for both parameters, setting one to 0 while testing the other against its corresponding detector. Figure~\ref{fig:adaptive} in Appendix~\ref{sec:appendix_adaptive} illustrates the fooling rates for the adaptive attack without detection and with ViTGuard-I and ViTGuard-II, respectively. We observe that, for CIFAR-100 and Tiny-Imagenet, setting $(\beta_{attn}, \beta_{cls})$ to $(10^{-2}$, $10^{-3})$ and $(10^{-3}$, $10^{-3})$, respectively, yields the highest fooling rates. Consequently, these optimal coefficients are set to evaluate the efficacy of our joint detectors against the adaptive attack.
Without ViTGuard, the adaptive attacks achieve decent fooling rates of 55.8\% for CIFAR-100 and 42.4\% for Tiny-ImageNet. However, ViTGuard reduces these fooling rates to 24.3\% and 11.6\% with an FPR of 0.01, and to 6.7\% and 5.6\% with an FPR of 0.05. Thus, the adaptive attacks fail to compromise ViTGuard.  Overall, despite attackers having full knowledge of the detection mechanism, generating successful adversarial samples remains a challenging task. Furthermore, the practical solution to the customized optimization problem of the adaptive attack is hindered by the non-deterministic image reconstruction process associated with the random masking strategy.

\section{Discussion and Future Work}

\textbf{Computational Overhead:}
 The overhead associated with our approach is largely due to the image reconstruction process, where the MAE model follows the architecture outlined in the original MAE work~\cite{masked_autoencoder}. In our testing environment, the average computation time for MAE is 2.4 ms, compared to 0.2 ms for ViT-16 during inference. This latency is acceptable given the significant robustness benefits. For datasets with smaller feature spaces, adopting a scaled-down version of MAE could further decrease the computational burden.  

 \noindent
 \textbf{ViT Variants:} The applications of transformers in the CV domain have been rapidly evolving. 
While this paper assesses the generalizability of the proposed detectors on DeiT and CaiT, the emergence of diverse ViT variants necessitates further investigation. Specifically, a thorough examination of their performance under various adversarial attacks and an evaluation of the effectiveness of existing defense methods on these variants are required.
Furthermore, the CLS token, inherited from transformer designs in NLP, is absent in some ViT variants, such as Swin~\cite{swin}. Therefore, future research should explore the adaptation of attention-based and CLS-based detectors to these different ViT variants.

 \noindent
\textbf{Broader Defensive Strategies:} To enhance the effectiveness and flexibility of ViTGuard, two advanced strategies can be considered in future research. (1) Ensemble voting: this strategy leverages MAE for image reconstruction and aggregates outputs from multiple detectors, both from our current research and existing studies. (2) Soft decision: this approach provides users with a confidence score, allowing for precise threshold adjustments to balance security and performance on clean samples.

\section{Conclusion}

This paper introduces ViTGuard as a novel approach for detecting adversarial examples of ViT classifiers. ViTGuard is designed for robust detection against both common $L_p$ norm attacks and more practical patch attacks.
In contrast, existing detection methods either cannot be adapted for ViT models or fail to achieve accurate detection across various types of attacks. The effectiveness of ViTGuard can be attributed to two key aspects: (1) MAE effectively mitigates the adversarial effect from adversarial images while minimizing reconstruction loss to normal images by randomly masking and recovering a fraction of patches; (2) Attention maps and CLS representations prove to be more effective in detection for ViT models than the model layers' outputs typically used by existing detection methods. The efficacy of ViTGuard is validated across various adversarial attacks and datasets, and the evaluation results demonstrate its superior detection performance on both $L_p$ norm attacks and patch attacks.



\bibliographystyle{ACM-Reference-Format}
\bibliography{references}

\appendix

\section{Attack Configurations} \label{appendix_attack}
The attack parameters are shown in Table~\ref{tab:attack_param}. In FGSM, PGD, APGD attacks, $\epsilon$ is the maximum perturbation, \#steps is the total number of iterations, $\epsilon_{step}$ is the perturbation added in each step, and $\rho$ is a parameter for step-size update. In CW attack, ``Confidence'' indicates the classification confidence on adversarial samples, and $lr$ is the learning rate. In Patch-Fool and Attention-Fool, \#patches refers to the number of perturbed patches, and $\alpha$ is the coefficient for the attention loss. In SGM, $\gamma$ is the decay factor to the gradient, and attacks on the surrogate model share the same parameters as PGD attacks. SE and TR employ DeiT-tiny as the source model and utilize the PGD attack.

\begin{table}[h]
\centering
\caption{Attack parameters.}
\label{tab:attack_param}
\resizebox{\columnwidth}{!}{
\begin{tabular}{|ll|}
\hline
\multicolumn{2}{|c|}{CIFAR-10 / CIFAR-100}                                                                   \\ \hline
\multicolumn{1}{|l|}{FGSM}       & $\epsilon = 0.03$                                            \\ \hline
\multicolumn{1}{|l|}{PGD}        & $\epsilon = 0.03, \epsilon_{step} = 0.003, \#steps = 10$     \\ \hline
\multicolumn{1}{|l|}{APGD}       & $\epsilon = 0.03, \#steps = 10, \rho = 0.75$                 \\ \hline
\multicolumn{1}{|l|}{CW}         & Confidence = 50, \#steps = 30, $lr = 0.01$                   \\ \hline
\multicolumn{1}{|l|}{Patch-Fool} & \multicolumn{1}{l|}{\begin{tabular}[c]{@{}l@{}} \#patches =  4 (ViT-16) and 1 (ViT-32), \\ $\alpha = 0.002$, \#steps  = 250, $lr = 0.05$\end{tabular}} \\ \hline
\multicolumn{1}{|l|}{Attention-Fool} & \multicolumn{1}{l|}{\begin{tabular}[c]{@{}l@{}} \#patches =  4 (ViT-16) and 1 (ViT-32), \\ Uses the same patch selection method as Patch-Fool\end{tabular}} \\ \hline
\multicolumn{1}{|l|}{SGM}         & \multicolumn{1}{l|}{\begin{tabular}[c]{@{}l@{}} Surrogate model: ResNet-18, $\gamma$=0.5; \\ same parameters as PGD\end{tabular}} \\ \hline
\multicolumn{1}{|l|}{SE}         & \multicolumn{1}{l|}{\begin{tabular}[c]{@{}l@{}} Surrogate model: DeiT-tiny; \\ PGD: $\epsilon = 0.03, \epsilon_{step} = 0.0006, \#steps = 50$ \end{tabular}} \\ \hline
\multicolumn{1}{|l|}{TR}         & \multicolumn{1}{l|}{\begin{tabular}[c]{@{}l@{}} Surrogate model: DeiT-tiny; \\ PGD: $\epsilon = 0.03, \epsilon_{step} = 0.0006, \#steps = 50$ \end{tabular}} \\ \hline
\multicolumn{2}{|c|}{Tiny-ImageNet}                                                             \\ \hline
\multicolumn{1}{|l|}{FGSM}       & $\epsilon = 0.06$                                                       \\ \hline
\multicolumn{1}{|l|}{PGD}        & $\epsilon = 0.06, \epsilon_{step} = 0.006, \#steps = 10$                                                             \\ \hline
\multicolumn{1}{|l|}{APGD}       & $\epsilon = 0.06, \#steps = 10, \rho = 0.75
$                                                              \\ \hline
\multicolumn{1}{|l|}{CW}         & Confidence =  50, \#steps  = 30, $lr = 0.02$                                                             \\ \hline
\multicolumn{1}{|l|}{Patch-Fool} & \multicolumn{1}{l|}{\begin{tabular}[c]{@{}l@{}} \#patches =  4 (ViT-16) and 1 (ViT-32), \\ $\alpha = 0.002$,  \#steps  = 250, $lr = 0.05$ \end{tabular}}                                                               \\ \hline
\multicolumn{1}{|l|}{Attention-Fool} & \multicolumn{1}{l|}{\begin{tabular}[c]{@{}l@{}} \#patches =  4 (ViT-16) and 1 (ViT-32), \\ Uses the same patch selection method as Patch-Fool \end{tabular}}                                                               \\ \hline
\multicolumn{1}{|l|}{SGM}         & \multicolumn{1}{l|}{\begin{tabular}[c]{@{}l@{}} Surrogate model: ResNet-18, $\gamma$=0.5, \\ same parameters as PGD\end{tabular}} \\ \hline
\multicolumn{1}{|l|}{SE}         & \multicolumn{1}{l|}{\begin{tabular}[c]{@{}l@{}} Surrogate model: DeiT-tiny; \\ PGD: $\epsilon = 0.06, \epsilon_{step} = 0.0012, \#steps = 50$ \end{tabular}} \\ \hline
\multicolumn{1}{|l|}{TR}         & \multicolumn{1}{l|}{\begin{tabular}[c]{@{}l@{}} Surrogate model: DeiT-tiny; \\ PGD: $\epsilon = 0.06, \epsilon_{step} = 0.0012, \#steps = 50$ \end{tabular}} \\ \hline
\end{tabular}
}
\end{table}

\section{Experimental Setting}
\subsection{Datasets} ~\label{appendix_dataset}
\textbf{CIFAR-10~\cite{cifar10}} consists of 60,000 color images labeled over ten object classes, with 6,000 images per class. 
The size of each image is $32 \times 32$. The dataset is divided into the training dataset with 50,000 images, the validation dataset with 3,000 images, and the testing dataset with 7,000 images. 
\textbf{CIFAR-100~\cite{cifar10}} includes 100 object classes. The image size and dataset size for training, testing, and validation are identical to those of CIFAR-10.
\textbf{Tiny-ImageNet~\cite{tiny-imagenet}} consists of 110,000 color images labeled over 200 classes. It is partitioned into three subsets: a training dataset containing 100,000 images, a validation dataset containing 3,000 images, and a testing dataset containing 7,000 images. The size of each image is $64 \times 64$.

\subsection{Settings of Baseline Detection Methods} \label{appendix_existing}
{\bf KD+BU}~\cite{KDBU}: The Gaussian bandwidth is 0.26, following the configuration in the original paper.
{\bf LID}~\cite{Lid}: The number of neighbors is set to 10 when calculating the distance distribution.
{\bf FS}~\cite{Feature_squeezing}: This method is composed of three image squeezers and a $L_1$ distance based detector. The settings of image squeezers are listed below:
(1) Decreasing color bit depth: the color depth is decreased to 7 bits.
    (2) Median smoothing: the filter size is $2 \times 2$.
    (3) Non-local smoothing: the search window size is 13, the patch size is 3, and the filter strength is 4.
 {\bf MSCN}~\cite{MSCN}: Following the methodology outlined in the original paper, we sample the MSCN coefficient histogram from -2 to 2 with a 0.05 interval, generating an 81-dimensional vector for each image.
{\bf DAE}~\cite{magnet}: We adopt the same denoising autoencoder architecture as outlined in the original paper. The autoencoder comprises three convolutional layers with kernels of sizes $3 \times 3$, $3 \times 3$, and $1 \times 1$, respectively, each followed by a sigmoid activation function. The autoencoder is trained using the clean training dataset with unit Gaussian noise at a volume of 0.025. 
{\bf GAN}~\cite{defense-gan}: The GAN network consists of a generator and a discriminator. The generator uses an encoder-decoder architecture to restore the image, with the encoder comprising four convolutional layers, each followed by batch normalization and ReLU activation. The discriminator classifies the generated image, featuring five convolutional layers and ending with a Sigmoid layer. The probability divergence is calculated using the Jensen-Shannon divergence, with the temperature set to 10.
{\bf ARMRO}~\cite{armro}: We select five patches with the highest attention scores and mask them using the input average. 
For the supervised methods, the adversarial algorithm used during the detector's training phase is consistent with the actual attack employed in the testing phase. To tailor KD+BU and LID for ViT models, we substitute the outputs of hidden convolutional layers with the outputs of transformer blocks.

\section{Supplemental results}
\subsection{Input Visualization}  \label{appendix:vis_inputs}
\begin{figure*}[t]
  \begin{minipage}[b]{\textwidth}
    \centering
    \includegraphics[width=\linewidth]{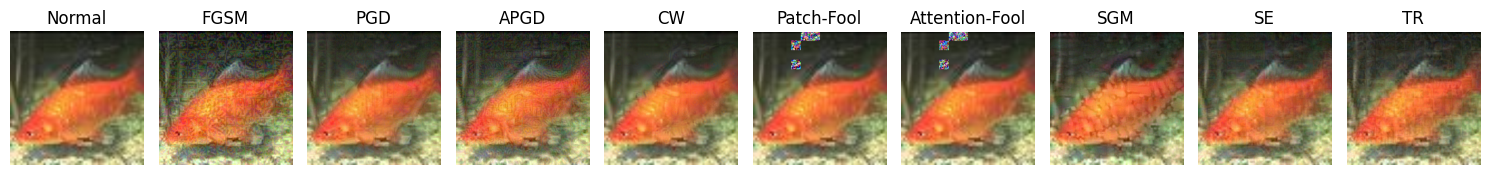}
    \vspace{-0.25in}
    \subcaption{Original images}
  \end{minipage}%
  \\
    \begin{minipage}[b]{\textwidth}
    \centering
    \includegraphics[width=\linewidth]{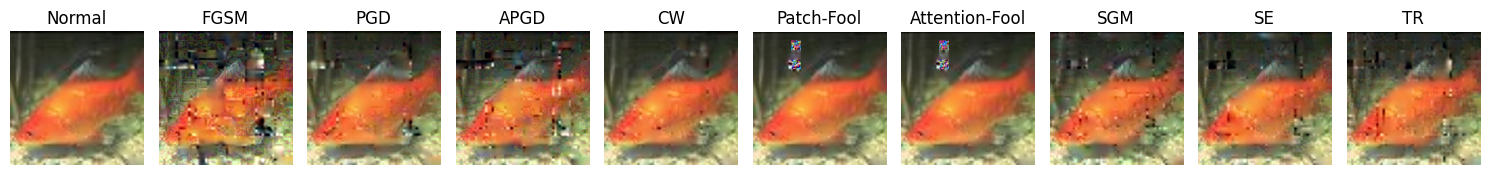}
    \vspace{-0.25in}
    \subcaption{Reconstructed images}
  \end{minipage}
  \caption{Visualization of original and reconstructed input images using the Tiny-ImageNet dataset. The original images include the normal image and its corresponding adversarial examples under different attacks, with the reconstructed images representing their versions reconstructed by the MAE model with a masking ratio of 0.5.}
  \label{fig:vis_inputs}
\end{figure*}

Figure~\ref{fig:vis_inputs} displays the normal input, adversarial inputs under various attacks, and the corresponding reconstructed images by the MAE model. 

\begin{figure*}[t]
  \begin{minipage}[b]{\textwidth}
    \centering
    \includegraphics[width=\linewidth]{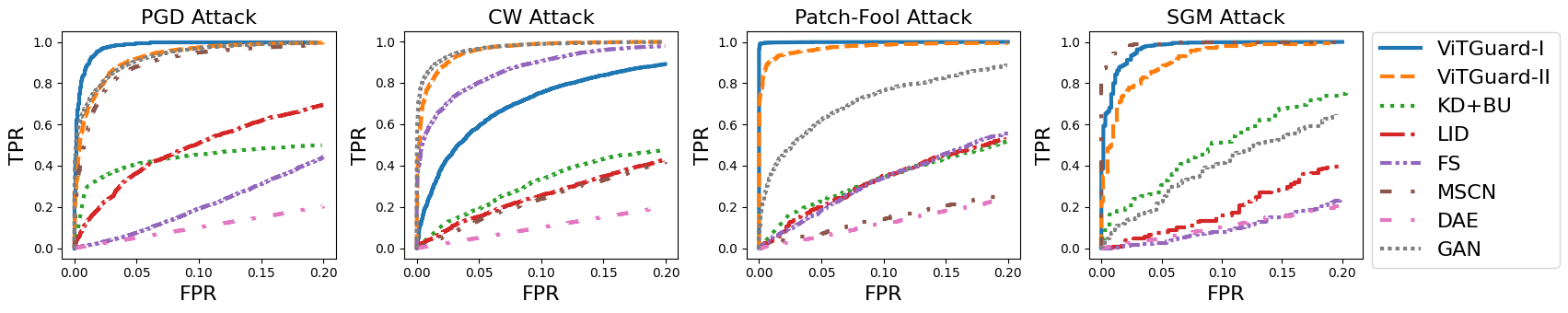}
    \vspace{-0.25in}
    \subcaption{CIFAR-10}
  \end{minipage}%
  \\
    \begin{minipage}[b]{\textwidth}
    \centering
    \includegraphics[width=\linewidth]{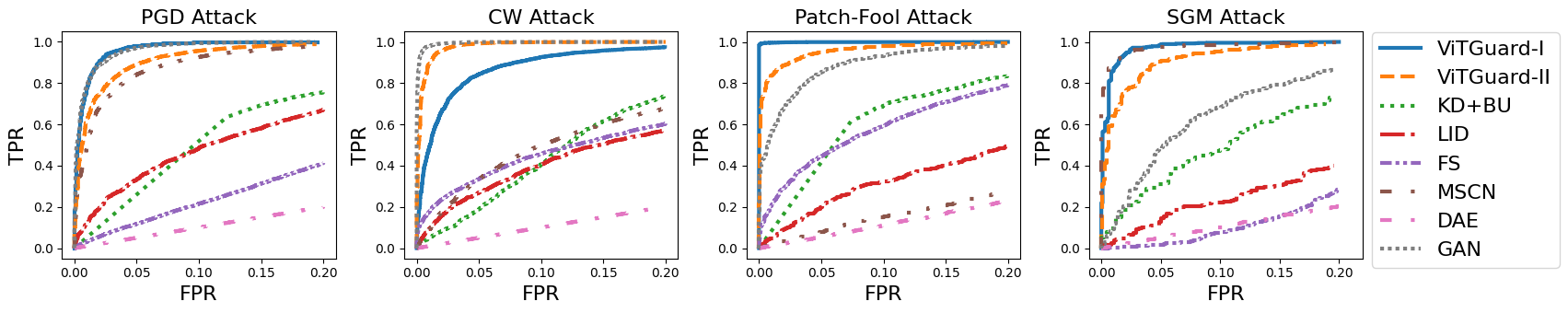}
    \vspace{-0.25in}
    \subcaption{CIFAR-100}
  \end{minipage}%
  \caption{ROC curves of different detection methods, including KD+BU, LID, FS, MSCN, DAE, GAN, and our individual detectors, applied to ViT-16 models.}
  \label{fig:roc_curves_cifar}
\end{figure*}

\begin{figure*}[t]
    \begin{minipage}[b]{\textwidth}
    \centering
    \includegraphics[width=\linewidth]{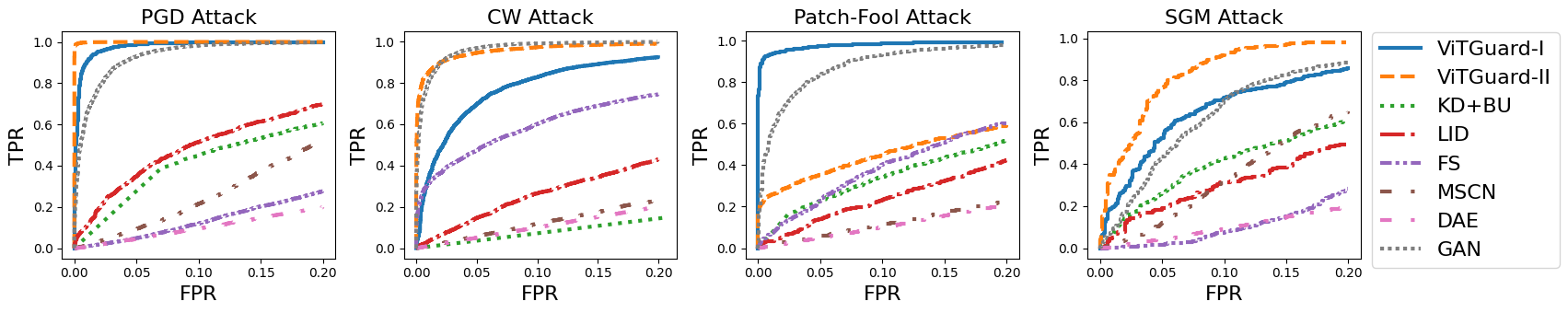}
  \end{minipage}
  \caption{ROC curves of different detection methods applied to ViT-16 models for Tiny-ImageNet.}
  \label{fig:roc_curves_tiny}
\end{figure*}

\subsection{Comparison between Half and Full Input Recovery} \label{sec:pre_half_entire}
To design an optimal strategy for deploying the MAE model that balances effectiveness with efficiency, we conducted preliminary experiments using the CIFAR-100 dataset and the ViT-16 model to compare detection performances for half and full-input recoveries. Full input recovery is achieved by applying the MAE model twice, targeting different areas, which can mitigate perturbations across the entire image. 
The performance comparison is shown in Figure~\ref{fig:half_full_recovery}.
Our findings indicate that fully recovering the input consistently leads to improved detection performance, with AUC scores enhancing by a range of 0.0001 to 0.007 for the more effective of the two detectors. However, this improvement is minimal, especially considering that recovering the full input incurs a larger computational overhead. Thus, we decided to apply the MAE model once to recover half of the input in the comprehensive evaluations.

\begin{table}[]
\centering
\caption{AUC scores for ViTGuard detectors on CIFAR-100: `Half' for half input recovery, `Entire' for full input recovery.}
\begin{tabular}{|c|cc|cc|}
\hline
\multirow{2}{*}{Attack} & \multicolumn{2}{c|}{ViTGuard-I}       & \multicolumn{2}{c|}{ViTGuard-II}      \\ \cline{2-5} 
                        & \multicolumn{1}{c|}{Half}    & Entire & \multicolumn{1}{c|}{Half}    & Entire \\ \hline
FGSM                    & \multicolumn{1}{c|}{0.9999}  & 1.0000 & \multicolumn{1}{c|}{0.9933}  & 0.9980 \\ \hline
PGD                     & \multicolumn{1}{c|}{0.9920}  & 0.9990 & \multicolumn{1}{c|}{0.9794}  & 0.9984 \\ \hline
APGD                    & \multicolumn{1}{c|}{0.9996}  & 1.0000 & \multicolumn{1}{c|}{0.9901}  & 0.9995 \\ \hline
CW                      & \multicolumn{1}{c|}{0.9694}  & 0.9899 & \multicolumn{1}{c|}{0.9956}  & 0.9989 \\ \hline
SGM                     & \multicolumn{1}{c|}{0.9939} & 0.9957 & \multicolumn{1}{c|}{0.9801} & 0.9882 \\ \hline
SE                      & \multicolumn{1}{c|}{0.9926}  & 0.9944 & \multicolumn{1}{c|}{0.9788}  & 0.9865 \\ \hline
TR                      & \multicolumn{1}{c|}{0.9957}  & 0.9969 & \multicolumn{1}{c|}{0.9909}  & 0.9951 \\ \hline
\end{tabular}
\label{fig:half_full_recovery}
\end{table}

\subsection{ROC Curves for Individual Detectors} \label{appendix_roc}

Figures~\ref{fig:roc_curves_cifar} and~\ref{fig:roc_curves_tiny} display ROC curves for various detection methods applied to ViT-16 models using the CIFAR-10, CIFAR-100, and Tiny-ImageNet datasets.

\subsection{Performance of Joint Detectors} \label{appendix_tpr_fpr}
Table~\ref{tab:tpr_fpr_cifar} shows the detection performance of our joint detectors under different FPR limits.

\begin{table}[t]
  \centering
  \caption{Performance of our joint detectors under different FPRs. For example, TPR (FPR@0.01) represents the TPR value under an FPR of 0.01. }
  \label{tab:tpr_fpr_cifar}
\resizebox{\columnwidth}{!}{
\begin{tabular}{|c|c|cc|cc|cc|}
\hline
\multirow{2}{*}{Attack} & \multirow{2}{*}{Metric} & \multicolumn{2}{c|}{CIFAR-10}         & \multicolumn{2}{c|}{CIFAR-100}        & \multicolumn{2}{c|}{Tiny-ImageNet}   \\ \cline{3-8} 
                        &                         & \multicolumn{1}{c|}{ViT-16} & ViT-32 & \multicolumn{1}{c|}{ViT-16} & ViT-32 & \multicolumn{1}{c|}{ViT-16} & ViT-32 \\ \hline

\multirow{2}{*}{FGSM}   & TPR (FPR@0.01)      & \multicolumn{1}{c|}{1.000}       &0.998        & \multicolumn{1}{c|}{1.000}       &0.994        & \multicolumn{1}{c|}{1.000}       &1.000        \\ \cline{2-8} 

& TPR (FPR@0.05)   & \multicolumn{1}{c|}{1.000}       &1.000        & \multicolumn{1}{c|}{1.000}       &1.000        & \multicolumn{1}{c|}{1.000}       &1.000        \\ \hline

\multirow{2}{*}{PGD}   & TPR (FPR@0.01)      & \multicolumn{1}{c|}{0.956}       &0.895        & \multicolumn{1}{c|}{0.898}       &0.863        & \multicolumn{1}{c|}{0.999}       &0.999        \\ \cline{2-8} 

& TPR (FPR@0.05)   & \multicolumn{1}{c|}{0.997}       &0.996        & \multicolumn{1}{c|}{0.999}       &0.992        & \multicolumn{1}{c|}{1.000}       &1.000        \\ \hline

\multirow{2}{*}{APGD}   & TPR (FPR@0.01)      & \multicolumn{1}{c|}{0.998}       &0.997        & \multicolumn{1}{c|}{0.994}       &0.993        & \multicolumn{1}{c|}{0.996}       &0.993        \\ \cline{2-8} 

& TPR (FPR@0.05)   & \multicolumn{1}{c|}{0.999}       &0.998        & \multicolumn{1}{c|}{0.995}       &0.995        & \multicolumn{1}{c|}{0.996}       &0.994        \\ \hline

\multirow{2}{*}{CW}   & TPR (FPR@0.01)      & \multicolumn{1}{c|}{0.830}       &0.979        & \multicolumn{1}{c|}{0.935}       &0.970        & \multicolumn{1}{c|}{0.867}       &0.897        \\ \cline{2-8} 

& TPR (FPR@0.05)   & \multicolumn{1}{c|}{0.980}       &0.999        & \multicolumn{1}{c|}{0.998}       &0.999        & \multicolumn{1}{c|}{0.964}       &0.962        \\ \hline

\multirow{2}{*}{SGM}   & TPR (FPR@0.01)      & \multicolumn{1}{c|}{0.870}       &0.740        & \multicolumn{1}{c|}{0.925}       &0.654        & \multicolumn{1}{c|}{0.432}       &0.129        \\ \cline{2-8} 

& TPR (FPR@0.05)   & \multicolumn{1}{c|}{0.988}       &0.983        & \multicolumn{1}{c|}{0.998}       &0.951        & \multicolumn{1}{c|}{0.868}       &0.462        \\ \hline

\multirow{2}{*}{SE}   & TPR (FPR@0.01)      & \multicolumn{1}{c|}{1.000}       &0.977        & \multicolumn{1}{c|}{0.920}       &0.719        & \multicolumn{1}{c|}{0.836}       &0.440        \\ \cline{2-8} 

& TPR (FPR@0.05)   & \multicolumn{1}{c|}{1.000}       &1.000        & \multicolumn{1}{c|}{0.998}       &0.951        & \multicolumn{1}{c|}{0.980}       &0.806        \\ \hline

\multirow{2}{*}{TR}   & TPR (FPR@0.01)      & \multicolumn{1}{c|}{0.999}       &0.903        & \multicolumn{1}{c|}{0.968}       & 0.835       & \multicolumn{1}{c|}{0.882}       &0.507        \\ \cline{2-8} 

& TPR (FPR@0.05)   & \multicolumn{1}{c|}{1.000}       &0.992        & \multicolumn{1}{c|}{0.999}       &0.979        & \multicolumn{1}{c|}{0.989}       &0.874        \\ \hline

\multirow{2}{*}{Patch-F}   & TPR (FPR@0.01)      & \multicolumn{1}{c|}{0.997}       &0.985        & \multicolumn{1}{c|}{0.997}       &0.986        & \multicolumn{1}{c|}{0.934}       &0.718        \\ \cline{2-8} 

& TPR (FPR@0.05)   & \multicolumn{1}{c|}{1.000}       &0.995        & \multicolumn{1}{c|}{0.999}       &0.995        & \multicolumn{1}{c|}{0.973}       &0.837        \\ \hline
\end{tabular}
}
\end{table}

\subsection{Classification Performance of ViT Variants} \label{appendix_variants}
The target models utilized in our study are DeiT-Base-Distilled-Patch16-224~\cite{0deit-base} and CaiT-S24-224~\cite{cait-s24}. These target models are fine-tuned on the Tiny-ImageNet dataset, freezing the self-attention layers and adjusting only the classification head.
Due to the different designs of transformer layers in these ViT variants, the attention-based detectors are specifically developed based on the last self-attention layer. In addition, CLS-based detectors are developed using the CLS representation from the classification head.
Table~\ref{tab:acc_variants} presents the classification accuracy of these models under various adversarial attacks. The attack parameters are provided in Table~\ref{tab:attack_param}. The results clearly indicate that the performance of ViT variants is significantly affected by these attacks.

\begin{table}[t]
\caption{Classification accuracy of ViT variants under various adversarial attacks on the Tiny-ImageNet dataset.}
\label{tab:acc_variants}
\begin{tabular}{|c|c|c|c|c|c|}
\hline
Model & No attack & PGD  & CW   & SGM & Patch-Fool \\ \hline
DeiT  & 83.83\%   & 0.00\% & 0.00\% &32.81\%     & 2.48\%            \\ \hline
CaiT  &83.45\%  &0.00\%   &0.00\%   &35.30\%     &7.63\%            \\ \hline
\end{tabular}
\end{table}

\subsection{Adaptive Attacks} \label{sec:appendix_adaptive}

 The revised optimization problem is presented as follows:
\begin{align}
\min J_{\mathrm{CW}}(x, x_{adv})&+ 
\beta_{attn} \sum_l^L J_{\mathrm{ATTN}}^{(l)}(x_{adv}, x^\prime) \nonumber \\
&+\beta_{cls} \sum_l^L J_{\mathrm{CLS}}^{(l)}(x_{adv}, x^\prime),
\end{align}
where $J_{\mathrm{CW}}$ is the original CW loss, $L$ is the total number of transformer blocks, $J_{\mathrm{ATTN}}^{(l)}$ denotes the $L_2$ distance between attention maps from the $l$-th transformer block of the adversarial image and the reconstructed image, and $J_{\mathrm{CLS}}^{(l)}$ measures the distance between CLS representations.

\begin{figure}[t]
    \centering
    
    \begin{subfigure}[b]{\columnwidth} 
        \centering
        \includegraphics[width=\textwidth]{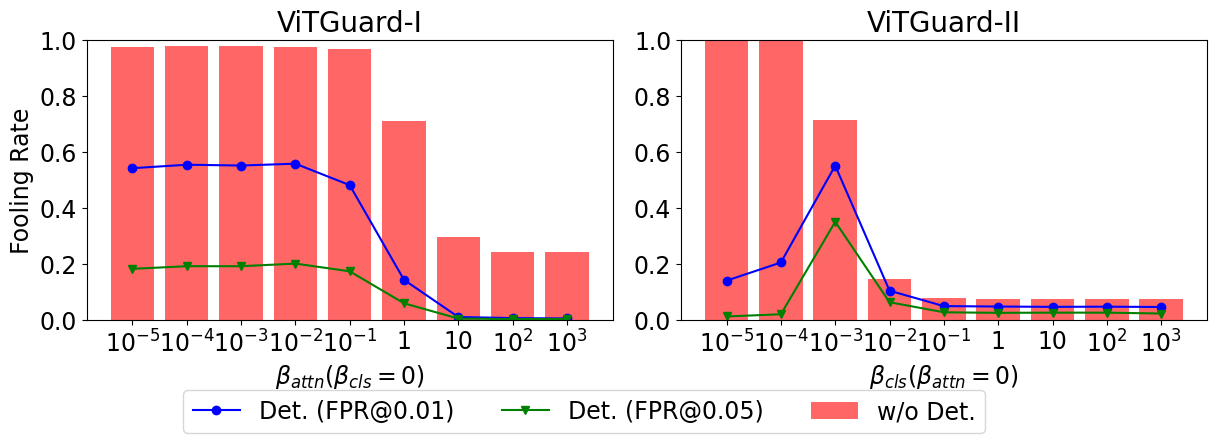}
        \caption{CIFAR-100}
        \label{fig:first_subfig}
    \end{subfigure}
    
    
    \begin{subfigure}[b]{\columnwidth} 
        \centering
        \includegraphics[width=\textwidth]{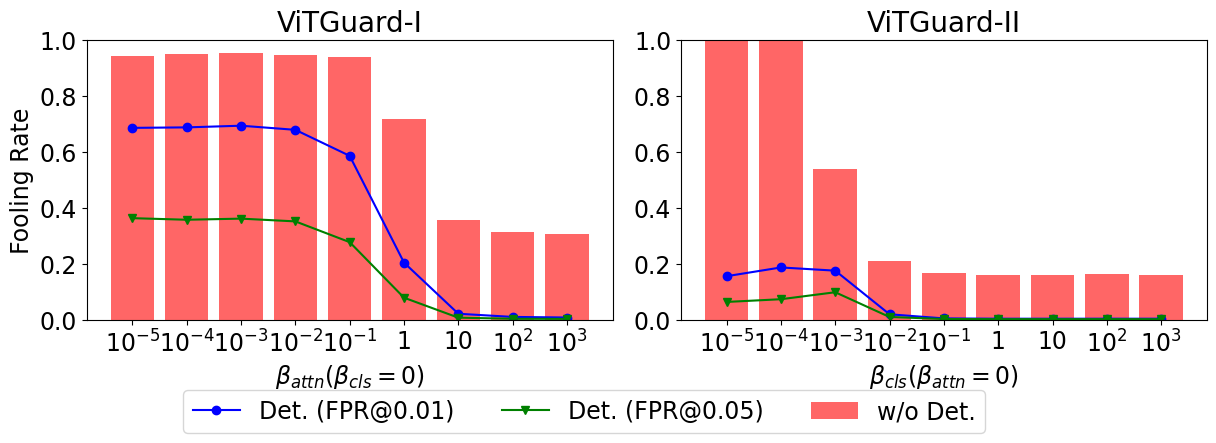}
        \caption{Tiny-ImageNet}
        \label{fig:second_subfig}
    \end{subfigure}
    
    \caption{Fooling rates of adaptive attacks without detection and with detection under different FPR limits.}
   \label{fig:adaptive}
\end{figure}

To determine the coefficients, namely $\beta_{attn}$ and $\beta_{cls}$, which balance the original CW loss with the attention distance and CLS representation distance, we set one coefficient to zero and vary the other within the range $[10^{-5}, 10^3]$. Subsequently, we analyze the fooling rates both without detection and with the corresponding individual ViTGuard detector. 
We consider three scenarios: (1) no detection, (2) ViTGuard with FPR limited to 0.01, and (3) ViTGuard with FPR limited to 0.05. 
 Figure~\ref{fig:adaptive} displays the fooling rates of adaptive attacks on the CIFAR-100 and Tiny-ImageNet datasets. To develop strong adaptive attacks, we select coefficients that allow the adaptive attack to achieve the highest fooling rate under detection. 
Eventually, for CIFAR-100 and Tiny-ImageNet, the values for $(\beta_{attn}, \beta_{cls})$ are set to $(10^{-2}$, $10^{-3})$ and $(10^{-3}$, $10^{-3})$ respectively.

\end{document}